\pdfoutput=1
\documentclass[11pt]{article}
\usepackage[table,dvipsnames]{xcolor}
\usepackage{collcell}
\usepackage{acl}

\usepackage{times}
\usepackage{latexsym}

\usepackage[T1]{fontenc}

\usepackage[utf8]{inputenc}

\usepackage{microtype}

\usepackage{tikz,pgfplots,pgfplotstable}
\usepackage{threeparttable}

\usepackage{colorblind}
\usepackage{soul}

\usepackage{inconsolata}

\usetikzlibrary{shapes.misc, positioning, decorations.pathreplacing, calc}
\usepackage{subcaption}

\usepackage{tcolorbox}
\definecolor{azure}{rgb}{0.0, 0.5, 1.0}
\tcbuselibrary{skins}
\tcbset{enhanced}
\newcommand{\qsecbox}[1]{\begin{tcolorbox}[left=1mm,right=1mm,boxrule=0.2mm,leftrule=2mm,drop fuzzy shadow,colframe=lightgray,frame style={left color=azure!90!lightgray}]#1\end{tcolorbox}}
\newcommand{\egcvalue}[1]{\textbf{\textit{#1}}}

\def\llm{77}

\usepackage{enumitem}
\usepackage{amssymb}
\usepackage{amsmath}
\usepackage{pgf}
\usepackage{float}
\usepackage{booktabs}
\usepackage{graphicx}
\usepackage{multirow}
\usepackage{float}
\usepackage{adjustbox}
\definecolor{Green}{RGB}{76, 119, 59}

\newcommand{\guillemet}[1]{``#1''}
\newcommand{\guillemettwo}[1]{\guillemotleft~#1~\guillemotright}

\title{QFrBLiMP: a Quebec-French Benchmark of Linguistic Minimal Pairs}

\author{David Beauchemin\thanks{\hspace{4pt}Contributed equally to this work.}~\hspace{1.5pt}\textsuperscript{\textdagger}, Pier-Luc Veilleux\protect\footnotemark[1]~\hspace{1.5pt}\textsuperscript{\textdaggerdbl}, Richard Khoury\textsuperscript{\textdagger} \and Johanna-Pascale Roy\textsuperscript{\textdaggerdbl}\\
Universit\'e Laval, Qu\'ebec, Canada \\
Computer Science Department\textsuperscript{\textdagger} and Department of Languages, Linguistics and Translation\textsuperscript{\textdaggerdbl}\\
\texttt{\{david.beauchemin, richard.khoury\}@ift.ulaval.ca} \\ \texttt{pier-luc.veilleux.1@ulaval.ca}\\
\texttt{johanna-pascale.roy@lli.ulaval.ca}
}

\newcommand{\johanne}[1]{\textcolor{red}{(J)}}

\usepackage{hyperref}
\usepackage[english]{babel}
\addto\extrasenglish{%

}

\usepackage{pgf}

\usepackage{collcell}
\usepackage{times}

\usepackage{expex}
\usepackage[linguistics]{forest}
\usepackage{extdash} 

\newcommand*{\MinNumber}{0.0}%
\newcommand*{\MidNumber}{60.0} %
\newcommand*{\MaxNumber}{100.0}%

\definecolor{Gray}{gray}{0.9}
\definecolor{cb-blue-green} {RGB}{ 0,  073,  073}
\definecolor{cb-green-sea}  {RGB}{ 0, 146, 146}
\definecolor{cb-rose}       {RGB}{255, 109, 182}
\definecolor{cb-salmon-pink}{RGB}{255, 182, 119}
\definecolor{cb-purple}     {RGB}{ 73,   0, 146}
\definecolor{cb-blue}       {RGB}{ 0, 109, 219}
\definecolor{cb-lilac}      {RGB}{182, 109, 255}
\definecolor{cb-blue-sky}   {RGB}{109, 182, 255}
\definecolor{cb-blue-light} {RGB}{182, 219, 255}
\definecolor{cb-burgundy}   {RGB}{146,   0,   0}
\definecolor{cb-brown}      {RGB}{146,  73,   0}
\definecolor{cb-clay}       {RGB}{219, 209,   0}
\definecolor{cb-green-lime} {RGB}{ 36, 255,  36}
\definecolor{cb-yellow}     {RGB}{255, 255, 109}
\definecolor{cb-grey}       {RGB}{233, 233, 233}

\newcommand{\ApplyGradient}[1]{%
        \ifdim #1 pt > \MidNumber pt
            \pgfmathsetmacro{\PercentColor}{max(min(100.0*(#1 - \MidNumber)/(\MaxNumber-\MidNumber),100.0),0.00)} %
            \hspace{-0.33em}\colorbox{SeaGreen!\PercentColor!Goldenrod!50}{#1}
        \else
            \pgfmathsetmacro{\PercentColor}{max(min(100.0*(\MidNumber - #1)/(\MidNumber-\MinNumber),100.0),0.00)} %
            \hspace{-0.33em}\colorbox{Red!\PercentColor!Goldenrod!50}{#1}
        \fi
}

\newcommand*{\MinNumberTwo}{-70}%
\newcommand*{\MidNumberTwo}{-30} %
\newcommand*{\MaxNumberTwo}{20}%

\newcommand{\ApplyGradienttwo}[1]{%
        \ifdim #1 pt > 0 pt
            \pgfmathsetmacro{\PercentColor}{max(min(100.0*(#1 - \MidNumberTwo)/(\MaxNumberTwo-\MidNumberTwo),100.0),0.00)} %
            \hspace{-0.33em}\colorbox{SeaGreen!\PercentColor!Goldenrod!50}{#1}
        \else
            \pgfmathsetmacro{\PercentColor}{max(min(100.0*(\MidNumberTwo - #1)/(\MidNumberTwo-\MinNumberTwo),100.0),0.00)} %
            \hspace{-0.33em}\colorbox{Red!\PercentColor!Goldenrod!50}{#1}
        \fi
}

\newcolumntype{R}{>{\collectcell\ApplyGradient}c<{\endcollectcell}}
\newcolumntype{D}{>{\collectcell\ApplyGradienttwo}c<{\endcollectcell}}

\begin{document}
\maketitle
\begin{abstract}
In this paper, we introduce the Quebec-French Benchmark of Linguistic Minimal Pairs (QFrBLiMP), a corpus designed to evaluate LLMs' linguistic knowledge of prominent grammatical phenomena in Quebec-French. 
QFrBLiMP comprises 1,761 minimal pairs annotated with 20 LPs.
Specifically, these minimal pairs have been created by manually modifying sentences extracted from an official online resource maintained by a Québec government institution. 
Each pair is annotated by 12 Quebec-French native speakers, who select the sentence they consider grammatical from the two.
These annotations are used to compare the competency of LLMs with that of humans.
We evaluate different LLMs on QFrBLiMP and MultiBLiMP-Fr by observing the rate of higher probabilities assigned to the sentences of each minimal pair for each category. 
We find that while grammatical competence scales with model size, a clear hierarchy of difficulty emerges. All benchmarked models consistently fail on phenomena requiring deep semantic understanding, revealing a critical limitation. 
Finally, our statistical analysis comparing QFrBLiMP and MultiBLiMP reveals a significant performance degradation for most models on Quebec-French; however, the most capable models remain within the statistical significance interval, demonstrating cross-dialectal robustness.
\end{abstract}

\section{Introduction}
\label{sec:intro}

Recent advancements in large language models (LLM) have demonstrated strong performance on a variety of Natural Language Processing (NLP) tasks, substantially increasing the performance of most NLP tasks \citep{zhang2023instruction, qin2024large}, such as insurance question-answering \citep{beauchemin2024quebec} and health applications \citep{bedi2025testing}.
LLMs were initially introduced for English \citep{kenton2019bert, brown2020language}, but many other languages were later introduced, such as Russian \citep{kuratov2019adaptation}, and French \citep{martin2020camembert}.
NLP research has approached the competencies evaluation of various natural language tasks of LLM with various benchmark corpora such as the English benchmarks GLUE \citep{wang-etal-2018-glue}, and BLiMP \citep{warstadt2020blimp} to name a few.
These corpora are collections of resources for training, evaluating, and analyzing LLMs \citep{eval-harness, chang2023survey}.
For example, GLUE aims to benchmark an NLP system's capabilities for natural language understanding (NLU) \citep{wang-etal-2018-glue}. 
At the same time, BLiMP focuses on evaluating the linguistic knowledge of LLMs on major grammatical phenomena in English.

However, a thorough understanding of how these models internalize linguistic knowledge is still developing. 
To this end, researchers have developed linguistic benchmarks to assess the linguistic competency (LC) of LLMs.
\citet{warstadt2020blimp} has introduced the benchmark of linguistic minimal pairs (LMP), which assesses the competency of LMs to acceptability contrasts using a pair of sentences where one is properly written and a minimal element is substituted in the second to induce a specific type of grammatical error \citep{chomsky2014aspects}.
Using this pair, one can evaluate whether an LM assigns a higher per-token probability to the acceptable sentence in each minimal pair (MP). 
This probability is held to indicate whether the model favors the grammatical sentence over the ungrammatical one and to determine which grammatical distinctions it is most sensitive to.
Such a benchmark provides indirect evidence about each model’s linguistic competence and enables fine-grained comparisons between them.
To benchmark LLM LC, \citet{warstadt2020blimp} also introduces human evaluation.
Namely, humans are presented with a MP and must select the one they feel is grammatical. 
Such human annotation helps compare the performance of LLM competency with that of humans.
Similar non-English benchmarks have been proposed to answer LLM LC in typologically diverse languages such as Japanese \citep{someya2023jblimp} and Dutch \citep{suijkerbuijk2024blimp}.
However, the competence of LLMs has not been tested in Quebec-French, which is the variety of French used in the Quebec provincial territory and taught in schools, prompted by governmental institutions.

To this end, we introduce the \textbf{Q}uebec-\textbf{Fr}ench \textbf{B}enchmark of \textbf{L}inguistic \textbf{M}inimal \textbf{P}airs (QFrBLiMP)\footnote{\href{https://github.com/davebulaval/QFrBLiMP}{https://github.com/davebulaval/QFrBLiMP}}, a corpus consisting of 1,761 MPs annotated with 20 linguistic phenomena (LP) and human evaluations.
Specifically, those MPs have been created by modifying sentences manually extracted from an official online resource maintained by a \textit{Québec} government institution. 
All sentences were classified by a linguist and annotated by twelve native French speakers from Quebec at the undergraduate level.
The main contributions of this work are twofold:
\begin{enumerate}[leftmargin=*, noitemsep, topsep=0ex]
    \item The creation and release of QFrBLiMP, a benchmark of LMP that leverage human-annotated sentences rather than synthetic ones;
    \item A set of experiments to assess the performance of LLM on QFrBLiMP as compared to that of human annotators.
\end{enumerate}

It is outlined as follows: first, we study the available LMP resources corpora in \autoref{sec:rel_work}. 
Then, we propose the QFrBLiMP corpus in \autoref{sec:QFrBLiMP}. In \autoref{sec:experiment} and \autoref{sec:res}, we present a set of experiments conducted on LLMs to test their competency in French. 
Finally, in \autoref{sec:conclusion}, we conclude and discuss our future work.

\section{Related Work}
\label{sec:rel_work}

\subsection{LLM Evaluation}
The evaluation of language models has historically been approached through two primary methodologies: automated metrics and performance on benchmark corpora. 
The first approach uses either task-agnostic metrics, like perplexity \citep{jelinek1977perplexity}, which measures how well a model predicts a given text sample, or task-specific metrics, such as the BLEU score \citep{Papineni02bleu:a} for assessing machine translation quality.

The second approach relies on large-scale benchmark corpora designed for downstream NLU or Natural Language Generation (NLG) tasks. 
For instance, the GLUE benchmark \citep{wang-etal-2018-glue} is used to evaluate a model's NLU performance on tasks including semantic similarity, sentiment analysis, and linguistic acceptability judgments. 
In contrast, the GLGE benchmark \citep{liu2021glge} focuses on NLG tasks like summarization and question answering. 
While informative, these benchmarks often test for functional linguistic competence that requires world knowledge and understanding and do not always disentangle it from a model's formal linguistic competence—its \guillemet{knowledge of rules and statistical regularities of a language} \citep{mcintosh2025inadequacies}. 
Moreover, because benchmarks like GLUE heavily adapt LLMs for specific downstream tasks, the evaluation does not necessarily reflect the knowledge that is inherently present in the pretrained model itself \citep{reuel2024betterbench, mcintosh2025inadequacies}.

\subsection{LLM Linguistic Acceptability Judgments Evaluation}
The evaluation of linguistic knowledge in LLMs often utilizes targeted syntactic evaluations.
Two approaches are used to perform this evaluation: binary classification acceptability judgments and MPs
\citep{warstadt2020blimp, chang2024survey, guo2024benchmarking, goyal2025iolbench}. 

In the first approach, a set of sentences that are either grammatical or ungrammatical, such as the two shown in \autoref{tab:itacola}, are provided to an LLM which must perform a binary classification \citep{warstadt2019neural}.
Eight corpora have been proposed to assess LLMs' capabilities to discriminate proper grammar from improper in their respective languages: CoLA for English \citep{warstadt2019neural}, DaLAJ for Swedish \citep{volodina2021dalaj}, ITACoLA for Italian \citep{trotta2021monolingual}, RuCoLA for Russian \citep{mikhailov2022rucola}, CoLAC for Chinese \citep{hu2023revisiting}, NoCoLA for Norwegian \citep{jentoft2023nocola}, JCoLa for Japanese \citep{someya2023jcola}, 
GalCoLA for Galician \citep{de2023dependency}, EsCoLA for spanish \citep{bel2024escola}, CatCoLA for Catalan \citep{bel2024catcola}, HuCoLA for Hungarian \citep{ligeti2024hulu}, and QFrCoLA for Quebec-French \citep{beauchemin2025qfrcola}.

The second approach is a method used in generative linguistics to evaluate human competence. 
It involves presenting pairs of sentences that are minimally different but contrast in grammatical acceptability \citep{warstadt2020blimp}. 
We present an example of a MP in \autoref{tab:blimp}.
An LLM is considered to have acquired a specific grammatical rule if it assigns a higher probability to the grammatical sentence than its ungrammatical counterpart.
Corpora such as BLiMP in English \citep{warstadt2019linguistic}, CLiMP \citep{xiang2021climp}, SLING \citep{song2022sling} and ZhoBLiMP \citep{liu2024zhoblimp} in Chinese, JBLiMP in Japanese \citep{someya2023jblimp}, BLiMP-NL in Dutch \citep{suijkerbuijk2024blimp}, RuBLiMP in Russian \citep{taktasheva2024rublimp}, BL2MP in Basque \citep{urbizu2024well}, and MultiBLiMP in 106 languages \citep{jumelet2025multiblimp} have been proposed to enable the evaluation of LM on a wide range of LP.
However, to date, no such corpus exists for Quebec-French\footnote{MultiBLiMP introduces French MPs, taken from a French resource. French in Quebec differs from France \citep{fagyal2006french}. For example, the feminization of titles differs between the two; the feminization of \textit{auteur} (author) in Quebec is accepted as \textit{autrice} or \textit{auteure} \citep{auteur}. In contrast, in France it is only accepted as \textit{auteure} \citep{auteuraf}.
However, both countries have similar LP, such as syntax and plurals \citep{dankova2017storytelling}.}.

\begin{table}
    \centering
    \resizebox{0.49\textwidth}{!}{%
    \begin{tabular}{cl}
    \toprule
    Label & \multicolumn{1}{c}{Sentence}           \\\midrule
    \texttt{0} (Ungrammatical)    & Edoardo returned to his last year city \\
    \texttt{1}  (Grammatical)   & This woman has impressed me          \\\bottomrule 
    \end{tabular}%
    }
    \caption{Example sentences from the ItaCoLA dataset \citep{trotta2021monolingual}.}
    \label{tab:itacola}
    \vspace{-1em}
\end{table}

\begin{table}
    \centering
    \footnotesize
    \begin{tabular}{l|l}
    Acceptable Sentence & Not Acceptable Sentence         \\\midrule
    The cats annoy Tim.     & The cats annoys Tim.
    \end{tabular}
    \caption{Example of a MP \citep{warstadt2019linguistic}.}
    \label{tab:blimp}
    \vspace{-2em}
\end{table}

\section{QFrBLiMP: \textbf{Q}uebec-\textbf{Fr}ench \textbf{B}enchmark of \textbf{L}inguistic \textbf{M}inimal \textbf{P}airs}
\label{sec:QFrBLiMP}
In this work, we introduce the \textbf{Q}uebec-\textbf{Fr}ench \textbf{B}enchmark of \textbf{L}inguistic \textbf{M}inimal \textbf{P}airs (QFrBLiMP), which will be the first large-scale MPs dataset, with human annotations, for the Quebec-French language.

\subsection{Sources}
\begin{figure}
    \centering
    \includegraphics[width=\linewidth]{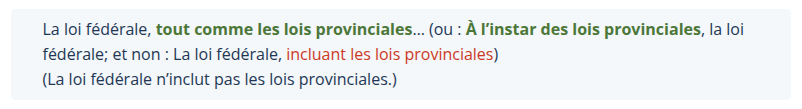}
    \caption{Snipped of the translated BDL article for present participles \guillemet{including} and \guillemet{excluding}.}
    \label{fig:bdl}
    \vspace{-1.5em}
\end{figure}

The QFrBLiMP dataset is composed of 1,761 Quebec-French MP sentences extracted from a prescriptivist source, the \guillemet{\textit{Banque de dépannage linguistique}} (BDL), an official online linguistic resource created by the Office québécois de la langue française (OQLF), a public organization of the province of Quebec (Canada). 
The BDL contains 2,667 articles organized into eleven categories, such as \guillemet{spelling} and \guillemet{syntax}. 
These articles explain various LP that the OQLF deems normatively correct or incorrect, using examples written by French linguists to illustrate each case based on linguistic observations.
For example, within the \guillemet{syntax} category, an article addresses the proper and improper use of the 
present participles \guillemet{including} and \guillemet{excluding}.
As shown in \autoref{fig:bdl}, the BDL provides examples of correct sentences (marked in \textcolor{Green}{\textbf{green}}) and an erroneous usage (marked in \textcolor{red}{\textbf{red}}) to illustrate normative use.
The source is publicly available online, and we obtained authorization to publish under a CC-BY-NC-SA 4.0 license.

\subsection{Data Collection}
Sentences in QFrBLiMP were manually collected from the BDL online resource. 
Specifically, we examined all 2,667 of its articles and extracted 25,153 linguistic acceptability judgment sentences.
Each sentence was labelled \texttt{0} (ungrammatical) or \texttt{1} (grammatical) following the BDL \textcolor{Green}{\textbf{green}}/\textcolor{red}{\textbf{red}} colour scheme as illustrated in \autoref{fig:bdl}, and MPs are manually organized following the colour scheme,  and classified into one of 20 LP \citep{fagyal2006french, chesley2010lexical, boivin2020categorisation, feldhausen2021revisiting}.
We present our twenty-category statistics in \autoref{tab:categorization_complete}, and we present examples and descriptions in \autoref{ann:phenomena}.

\begin{table*}
    \centering
    \tiny
    \begin{tabular}{lclc}
        \toprule
        Linguistic Phenomena & \# Pairs & Linguistic Phenomena & \# Pairs\\
        \midrule
        \texttt{1} \textit{Accords participes passés}
        (Past participle agreements)      
        & 97 & \texttt{11} \textit{Accord des adjectifs} (Adjective agreement)      
        & 100\\
        \texttt{2} \textit{Flexion du verbe} (Verb inflection)        
        & 95 &         \texttt{12} \textit{-é / -er}        
        & 100\\
        \texttt{3} \textit{ne ... que} (only ... that)      
        & 97 & \texttt{13} \textit{Sélection lexicale du complément}
        (Lexical selection of the complement)      
        & 96\\
        \texttt{4} \textit{Sélection morphologie fonctionnelle}
        (Functional morphology selection)       
        & 96 &
        \texttt{14} \textit{Négation de l'infinitif}        (Infinitive negation)       
        & 97\\
        \texttt{5} \textit{Clitique dans la négation de l'infinitif}
        (Clitics in infinitive negation)     
        & 99 &         \texttt{15} \textit{Îlot sujet}
        (Subject island)       
        & 60\\
        \texttt{6} \textit{Montée du clitique}
        (Rising clitics)       
        & 97&         \texttt{16} \textit{Îlot ajout}
        (Addition island)        
        & 60\\
        \texttt{7} \textit{Négation standard}
        (Standard negation)        
        & 114&        \texttt{17} \textit{Îlot qu-}
        (Island qu-)       
        & 60\\
        \texttt{8} \textit{Déterminants}
        (Determinants)        
        & 106&        \texttt{18} \textit{Îlot SN}
        (SN island)       
        & 60\\
        \texttt{9} \textit{Sémantique lexicale}
        (Lexical semantics)       
        & 113&        \texttt{19} \textit{Dépendance parasitique avec dont}
        (Parasitic dependence with including)         
        & 60\\
        \texttt{10} \textit{Accord dans l'expression idiomatique}
        (Agreement in idiomatic expression)       
        & 91& \texttt{20} \textit{Préposition orpheline}
        (Orphan preposition)       
        & 63
        \\\bottomrule
    \end{tabular}
    \caption{QFrBLiMP LP and number of pairs (\# Pairs).}
    \label{tab:categorization_complete}
    \vspace{-1.75em}
\end{table*}

\subsubsection{Human Evaluation Methodology}
\label{subsubsec:automatic_eval}
We collect human judgment of our MPs. Following the arguments of \citet{van-der-lee-etal-2019-best}, we present our human evaluation methodology.

\paragraph{Number of outputs.} 
We randomly selected 1,761 MPs for annotation and 15 for practice\footnote{These practice annotations have been used to help annotators practice their tasks and adjust our guidelines; these examples have been discarded from the final dataset.}.

\paragraph{Presentation and interface.} 
We used a customized version of the Prodigy annotation tool \citep{prodigy_montani_honnibal}, and we present the interface in \autoref{ann:clf_annotation} (in French).
Annotators use our annotation procedure to annotate each instance randomly. 

\paragraph{Annotators.}
We selected twelve native Quebec-French-speaking students at the University Laval as our annotators.
A meeting was held with them to introduce the task, instructions, and annotation guide and interface.
Instructions included that they must spend at most 5 minutes per sentence pair.
Furthermore, the 15 practice instances were annotated during a pilot phase to familiarize them with the task.
Finally, during a second meeting after evaluating the practice instances, annotators received feedback and advice on what phenomena they should be cautious about.
Recognizing the significant contribution of our annotators, they were remunerated fairly according to the University's hourly salary pay scale.
Each annotator completed their work in at most 20 hours.
We provide in-depth details of the evaluation setup in our Human Evaluation Datasheet \citep{shimorina2021human} in \autoref{ann:hed}.

\paragraph{Annotation Procedure.}
The annotators are prompted with an MP, and they must select \textbf{one} sentence. The order in which the two sentences appear, either the grammatical or ungrammatical one first, is randomized to reduce the risk of an annotator always selecting the same response. 

\paragraph{Final Annotation.} To select the final label, we use a majority vote, and in case of ties, we randomly select one of the two options (2 occur.).

\subsection{Annotation Results}
\label{subsec:annotation_res}
We compute the inter-annotator agreement using the Worker Agreement With Aggregate (WAWA) coefficient \citep{ning-etal-2018-joint}, which indicates the average fraction of the annotators’ votes that agree with the aggregated vote for each pair.
We can see that a significant number of phenomena achieved high agreement rates, with several exceeding 90\%, peaking at 96.08\% for phenomenon LP 12 and averaging 86.31\% (not in the table).
These high scores suggest the annotation guidelines for these categories are clear and robust. In contrast, a few phenomena proved more challenging for annotators, as evidenced by significantly lower agreement for phenomenon LP 19 (72.78\%) and 20 (69.84\%). 
This variance indicates that certain phenomena may be inherently more subjective or require refined definitions to improve annotation consistency. Despite these outliers, the generally high agreement across the majority of phenomena confirms the reliability of the resulting dataset.

\begin{table}
    \centering
    \tiny
    \begin{tabular}{lclclclc}
    \toprule
        LP & \begin{tabular}[c]{@{}c@{}}WAWA\\(\%) ($\uparrow$)\end{tabular} & LP & \begin{tabular}[c]{@{}c@{}}WAWA\\(\%) ($\uparrow$)\end{tabular} & LP & \begin{tabular}[c]{@{}c@{}}WAWA\\(\%) ($\uparrow$)\end{tabular} & LP & \begin{tabular}[c]{@{}c@{}}WAWA\\(\%) ($\uparrow$)\end{tabular} \\
        \midrule
        \texttt{1} & 86.51 & \texttt{6} & 92.18 & \texttt{11} & 88.25 & \texttt{16} & 84.44 \\
        \texttt{2} & 90.09 & \texttt{7} & 91.30 & \texttt{12} & 96.08 & \texttt{17} & 83.75 \\
        \texttt{3} & 93.30 & \texttt{8} & 86.56 & \texttt{13} & 86.20 & \texttt{18} & 75.56 \\
        \texttt{4} & 93.32 & \texttt{9} & 78.69 & \texttt{14} & 95.79 & \texttt{19} & 72.78 \\
        \texttt{5} & 95.54 & \texttt{10} & 90.66 & \texttt{15} & 75.28 & \texttt{20} & 69.84 \\
        \bottomrule                       
        \end{tabular}
        \caption{Per-linguistic phenomena (LP) WAWA inter-annotator agreement rates (\%). $\uparrow$ means higher is better.}
        \label{tab:annotators_res}
    \vspace{-2em}
\end{table}

\subsection{Comparison With Other Similar Corpora}
We present our corpus statistics in \autoref{tab:stats}, along with other similar MP benchmarks for comparison. We can see that, while QFrBLiMP is smaller than synthetic large-scale corpora like BLiMP or MultiBLiMP, it offers a competitive 20 LPs, surpassing benchmarks like CLiMP (16) and providing significantly broader paradigmatic coverage than MultiBLiMP (6). 
Moreover, all pairs are drawn from an official Quebec-French grammar source, rather than a synthetic processing pipeline; ours was created from human annotation, ensuring a higher-quality corpus.
This grammar-driven approach ensures that the dataset precisely tests attested LP specific to this variety.
This provides a focused, rich dataset for evaluating models' grammatical understanding in Quebec-French. 
\begin{table}
    \centering
    \resizebox{0.48\textwidth}{!}{%
        \begin{tabular}{@{}lcccc@{}}
        \toprule
                                                                 & Language     &Is Synthetic & \# Pairs & LP \\ \midrule
        BLiMP \citep{warstadt2019linguistic}     & English   & Y   & 67,000           & 67                   \\
        CLiMP \citep{xiang2021climp}             & Chinese   & Y    & 16,000           & 16                   \\
        SLING \citep{song2022sling}                & Chinese   & Y    & 38,000           & 38                   \\
        ZhoBLiMP \citep{liu2024zhoblimp}         & Chinese    & Y   & 35,000           & 118                  \\
        JBLiMP \citep{someya2023jblimp}          & Japanese   & N   & 331              & 39                   \\
        BLiMP-NL \citep{suijkerbuijk2024blimp}   & Dutch     & Y    & 8,400            & 84                   \\
        RuBLiMP \citep{taktasheva2024rublimp}    & Russian    & Y   & 45,000           & 45                   \\
        MutliBLiMP \citep{jumelet2025multiblimp} & 106 & Y & 125,000         & 6                 \\\midrule
        QFrBLiMP (ours)                                          & Quebec-French & N & 1,761            & 20                   \\ \bottomrule
        \end{tabular}%
    }
    \caption{Comparison of MPs benchmarks.}
    \label{tab:stats}
    \vspace{-1.5em}
\end{table}

\section{Experiments}
\label{sec:experiment}
We benchmark \llm{} open-source LLMs on QFrBLiMP and the Metropolitan French part of MultiBLiMP for comparison (255 instances).
We also benchmark these models against two baselines.
As our first baseline, \texttt{Random}, we randomly selected one of the two sentences, while our second, \texttt{human}, is the majority of our annotators' answers.

\subsection{Evaluation Metrics}
Following \citet{warstadt2020blimp}, performance is measured using the accuracy score.

\subsection{Method}
Following \citet{warstadt2020blimp, taktasheva2024rublimp}, the sentences in an MP are ranked based on their perplexity (PPL). 
The PPL of a sentence $s$ is inferred as \autoref{eq:ppl}, where $|s|$ is the sentence length in tokens, $x_i$ are individual words, and $\Theta$ denotes the LLM’s parameters. We use the official source code from \href{https://huggingface.co/spaces/zoebat20/BLiMP/blob/main/app.py}{BLiMP HuggingFace Metrics repository} to compute the probability.

\vspace{-13pt}
    \begin{equation}
      \text{PPL}(s) = \exp\left(-\frac{1}{|s|}\sum_{i=0}^{|s|} 
    \log\left({P_{\Theta}}\left(x_i|x_{<i}\right)\right)\right)
    \label{eq:ppl}
\end{equation}

\subsection{Selected Large Language Models}
To ensure a thorough and representative analysis of the current open-source LLM landscape, we selected \llm{} based on several criteria. Our selection aims to cover three aspects of LLM specifications:

\begin{enumerate}[leftmargin=*, noitemsep, topsep=0ex]
    \item \textbf{Variety in Size:} The selected models span a large range of parameter counts, from smaller models under 3 billion parameters to the largest open-source model we can fit on our hardware\footnote{We rely on three NVIDIA RTX 6000 ADA with 49 GB of memory, without memory pooling, thus the maximum size we can fit is around 72B parameters.} (details in \autoref{tab:selectedllm}).
    \item \textbf{Variety in Capability:} We intentionally included models marketed as having advanced \guillemet{reasoning} ($\Gamma$) capabilities to assess if this specialization translates to better performance on our knowledge-intensive task.
    \item \textbf{Instruction-Tuning}: We included models that have been instruction-tuned (\texttt{it}) to compare against their base model counterpart.
    \item \textbf{Model Specialization:} We included models with a declared specialization in French ($\Upsilon$), to test whether this focus provides an advantage.
\end{enumerate}

We present our selected model, model source, and size in \autoref{an:selectedllmdetails}.
Since we use perplexity to compute the selected sentence, we cannot benchmark private LLMs since we do not have access to the decoder probabilities.

\section{Results and Discussion}
\label{sec:res}

In this section, we present the comprehensive results of benchmarking our \llm{} selected LLMs for our experiments. 
Our findings are presented in two tables and a two summary figure, which collectively provide a multifaceted analysis of model performance. 
\autoref{tab:results} presents the best results raw per-phenomenon accuracy for each model against our \texttt{Random} and \texttt{Human} baselines, offering a direct measure of grammatical competence. We also present the complete results in \autoref{ann:completeresults}.
To provide a clearer perspective on these results, \autoref{tab:resultsdiff} reframes this performance as a differential against the \texttt{Human} baseline, immediately revealing which phenomena are mastered and which remain challenging relative to human judgment. 
We also present the complete results in \autoref{ann:completeresults}.
Complementing this granular, per-phenomenon analysis, \autoref{fig:tendancies} illustrates the overarching trend by plotting model accuracy as a function of scale, visualizing the strong positive correlation between LLM size and grammatical competency, while \autoref{fig:versusmulti} plots the performance on QFrBLiMP against MultiBLiMP for each model, revealing not only a strong correlation but also consistently higher scores on our benchmark. 
Finally, to further analyze the performance of Quebec-French and French Metropolitan, we compare each LLM's result over QFrBLiMP against its result on MultiBLiMP using a Z-test for statistical significance \cite{lawley1938generalization}. Our null hypothesis is that the pair of accuracies are equal, meaning that Z-test values outside $|3.290527|$ allow us to reject the hypothesis with $\alpha = 0.001$ (i.e. not a significant difference between the two benchmarks). 
A positive value indicates that French Metropolitan performs significantly better than Quebec French, and a negative value means the opposite.
We present in \autoref{fig:ztest} a visual representation of the test results of each model tested on the MultiBLiMP (x-axis) and QFrBLiMP (y-axis) benchmarks. 
To simplify the analysis, we separated the models into three groups based on their performance. 
In \autoref{fig:ztest}, models in red have a performance that is lower than our random-selection baseline (marked by the black dashed lines) on at least one of the two benchmarks. 

\begin{table*}
    \centering
    \resizebox{\textwidth}{!}{%
    \begin{tabular}{lRRRRRRRRRRRRRRRRRRRR}
        \toprule
        LLM & \multicolumn{1}{c}{\texttt{1}} & 
        \multicolumn{1}{c}{\texttt{2}} & 
        \multicolumn{1}{c}{\texttt{3}} & 
        \multicolumn{1}{c}{\texttt{4}} & 
        \multicolumn{1}{c}{\texttt{5}} & 
        \multicolumn{1}{c}{\texttt{6}} & 
        \multicolumn{1}{c}{\texttt{7}} & 
        \multicolumn{1}{c}{\texttt{8}} &
        \multicolumn{1}{c}{\texttt{9}} & 
        \multicolumn{1}{c}{\texttt{10}} & 
        \multicolumn{1}{c}{\texttt{11}} &
        \multicolumn{1}{c}{\texttt{12}} & 
        \multicolumn{1}{c}{\texttt{13}} &
        \multicolumn{1}{c}{\texttt{14}} &
        \multicolumn{1}{c}{\texttt{15}} & 
        \multicolumn{1}{c}{\texttt{16}} & 
        \multicolumn{1}{c}{\texttt{17}} &
        \multicolumn{1}{c}{\texttt{18}} &
        \multicolumn{1}{c}{\texttt{19}} &
        \multicolumn{1}{c}{\texttt{20}} \\
        \midrule
        \texttt{Bloom-1b7} & 88.66 & 97.89 & 97.94 & 92.71 & 96.97 & 94.85 & 91.23 & 85.85 & 69.03 & 93.41 & 91.00 & 100.00 & 80.21 & 97.94 & 86.67 & 96.67 & 80.00 & 81.67 & 83.33 & 28.57 \\
        \texttt{Bloom-560m} & 89.69 & 90.53 & 98.97 & 92.71 & 95.96 & 91.75 & 88.60 & 85.85 & 61.95 & 91.21 & 90.00 & 97.00 & 84.38 & 97.94 & 81.67 & 90.00 & 76.67 & 76.67 & 80.00 & 20.63 \\
        \texttt{CamemBERT-large} ($\Upsilon$) & 48.45 & 38.95 & 42.27 & 62.50 & 30.30 & 49.48 & 82.46 & 55.66 & 40.71 & 52.75 & 65.00 & 52.00 & 35.42 & 22.68 & 93.33 & 100.00 & 90.00 & 95.00 & 65.00 & 0.00 \\
        \texttt{Chocolatine-2-14b-it} ($\Upsilon$) & 91.75 & 98.95 & 95.88 & 97.92 & 96.97 & 96.91 & 95.61 & 88.68 & 71.68 & 92.31 & 90.00 & 98.00 & 84.38 & 89.69 & 83.33 & 95.00 & 81.67 & 81.67 & 73.33 & 42.86 \\
        \texttt{Gemma-2-9b} ($\Gamma$) & 88.66 & 95.79 & 94.85 & 95.83 & 97.98 & 92.78 & 94.74 & 89.62 & 67.26 & 93.41 & 90.00 & 98.00 & 81.25 & 81.44 & 98.33 & 100.00 & 95.00 & 91.67 & 81.67 & 30.16 \\
        \texttt{Lucie-7b} ($\Upsilon$) & 89.69 & 98.95 & 95.88 & 96.88 & 98.99 & 94.85 & 96.49 & 92.45 & 66.37 & 96.70 & 94.00 & 100.00 & 83.33 & 89.69 & 90.00 & 98.33 & 93.33 & 95.00 & 75.00 & 39.68 \\
        \texttt{Meta-Llama-$3.1$-70b-it} ($\Gamma$) & 91.75 & 94.74 & 93.81 & 95.83 & 96.97 & 95.88 & 89.47 & 88.68 & 70.80 & 91.21 & 95.00 & 93.00 & 83.33 & 90.72 & 95.00 & 100.00 & 88.33 & 91.67 & 73.33 & 41.27 \\
        \texttt{Meta-Llama-$3.1$-70b} ($\Gamma$) & 91.75 & 96.84 & 94.85 & 93.75 & 97.98 & 94.85 & 94.74 & 94.34 & 75.22 & 93.41 & 92.00 & 96.00 & 82.29 & 86.60 & 96.67 & 100.00 & 86.67 & 86.67 & 68.33 & 49.21 \\
        \texttt{Qwen$2.5$-14b} & 93.81 & 97.89 & 98.97 & 96.88 & 95.96 & 95.88 & 95.61 & 90.57 & 69.91 & 94.51 & 89.00 & 97.00 & 84.38 & 90.72 & 86.67 & 95.00 & 85.00 & 86.67 & 78.33 & 33.33 \\
        \texttt{Qwen$2.5$-72b} & 93.81 & 98.95 & 94.85 & 97.92 & 98.99 & 95.88 & 96.49 & 89.62 & 64.60 & 95.60 & 94.00 & 99.00 & 83.33 & 89.69 & 90.00 & 98.33 & 91.67 & 90.00 & 78.33 & 60.32 \\
        \texttt{Qwen$2.5$-7b} & 94.85 & 97.89 & 94.85 & 97.92 & 98.99 & 96.91 & 95.61 & 85.85 & 69.03 & 92.31 & 93.00 & 98.00 & 82.29 & 90.72 & 93.33 & 96.67 & 91.67 & 85.00 & 81.67 & 42.86 \\
        \midrule
        \texttt{Random} & 51.55 & 49.47 & 42.27 & 53.12 & 45.45 & 37.11 & 54.39 & 53.77 & 53.10 & 47.25 & 54.00 & 55.00 & 50.00 & 49.48 & 41.67 & 51.67 & 40.00 & 53.33 & 43.33 & 55.56 \\
        \texttt{Human} & 90.72 & 90.53 & 94.85 & 90.62 & 91.92 & 92.78 & 92.98 & 89.62 & 81.42 & 86.81 & 94.00 & 93.00 & 80.21 & 98.97 & 75.00 & 90.00 & 95.00 & 73.33 & 76.67 & 84.13 \\
        \bottomrule
    \end{tabular}
    }
    \captionsetup[figure]{font=small,labelfont=small}
    \caption{Subset of the average accuracy scores (\%) (higher is better), by phenomenon, of the LLMs and our two baselines. \guillemet{$\Upsilon$} are model that have a specialization in French, while \guillemet{$\Gamma$} are reasoning one.}
    \label{tab:results}
    \vspace{-1em}
\end{table*}

\begin{table*}
    \centering
    \resizebox{\textwidth}{!}{%
    \begin{tabular}{lDDDDDDDDDDDDDDDDDDDD}
        \toprule
        LLM & \multicolumn{1}{c}{\texttt{1}} & 
        \multicolumn{1}{c}{\texttt{2}} & 
        \multicolumn{1}{c}{\texttt{3}} & 
        \multicolumn{1}{c}{\texttt{4}} & 
        \multicolumn{1}{c}{\texttt{5}} & 
        \multicolumn{1}{c}{\texttt{6}} & 
        \multicolumn{1}{c}{\texttt{7}} & 
        \multicolumn{1}{c}{\texttt{8}} &
        \multicolumn{1}{c}{\texttt{9}} & 
        \multicolumn{1}{c}{\texttt{10}} & 
        \multicolumn{1}{c}{\texttt{11}} &
        \multicolumn{1}{c}{\texttt{12}} & 
        \multicolumn{1}{c}{\texttt{13}} &
        \multicolumn{1}{c}{\texttt{14}} &
        \multicolumn{1}{c}{\texttt{15}} & 
        \multicolumn{1}{c}{\texttt{16}} & 
        \multicolumn{1}{c}{\texttt{17}} &
        \multicolumn{1}{c}{\texttt{18}} &
        \multicolumn{1}{c}{\texttt{19}} &
        \multicolumn{1}{c}{\texttt{20}} \\
        \midrule
        \texttt{Bloom-1b7} & -2.06 & 7.36 & 3.09 & 2.09 & 5.05 & 2.07 & -1.75 & -3.77 & -12.39 & 6.60 & -3.00 & 7.00 & 0.00 & -1.03 & 11.67 & 6.67 & -15.00 & 8.34 & 6.66 & -55.56 \\
        \texttt{Bloom-560m} & -1.03 & 0.00 & 4.12 & 2.09 & 4.04 & -1.03 & -4.38 & -3.77 & -19.47 & 4.40 & -4.00 & 4.00 & 4.17 & -1.03 & 6.67 & 0.00 & -18.33 & 3.34 & 3.33 & -63.50 \\
        \texttt{CamemBERT-large} ($\Upsilon$) & -42.27 & -51.58 & -52.58 & -28.12 & -61.62 & -43.30 & -10.52 & -33.96 & -40.71 & -34.06 & -29.00 & -41.00 & -44.79 & -76.29 & 18.33 & 10.00 & -5.00 & 21.67 & -11.67 & -84.13 \\
        \texttt{Chocolatine-2-14b-it} ($\Upsilon$) & 1.03 & 8.42 & 1.03 & 7.30 & 5.05 & 4.13 & 2.63 & -0.94 & -9.74 & 5.50 & -4.00 & 5.00 & 4.17 & -9.28 & 8.33 & 5.00 & -13.33 & 8.34 & -3.34 & -41.27 \\
        \texttt{Gemma-2-9b} ($\Gamma$) & -2.06 & 5.26 & 0.00 & 5.21 & 6.06 & 0.00 & 1.76 & 0.00 & -14.16 & 6.60 & -4.00 & 5.00 & 1.04 & -17.53 & 23.33 & 10.00 & 0.00 & 18.34 & 5.00 & -53.97 \\
        \texttt{Lucie-7b} ($\Upsilon$) & -1.03 & 8.42 & 1.03 & 6.26 & 7.07 & 2.07 & 3.51 & 2.83 & -15.05 & 9.89 & 0.00 & 7.00 & 3.12 & -9.28 & 15.00 & 8.33 & -1.67 & 21.67 & -1.67 & -44.45 \\
        \texttt{Meta-Llama-$3.1$-70b-it} ($\Gamma$) & 1.03 & 4.21 & -1.04 & 5.21 & 5.05 & 3.10 & -3.51 & -0.94 & -10.62 & 4.40 & 1.00 & 0.00 & 3.12 & -8.25 & 20.00 & 10.00 & -6.67 & 18.34 & -3.34 & -42.86 \\
        \texttt{Meta-Llama-$3.1$-70b} ($\Gamma$) & 1.03 & 6.31 & 0.00 & 3.13 & 6.06 & 2.07 & 1.76 & 4.72 & -6.20 & 6.60 & -2.00 & 3.00 & 2.08 & -12.37 & 21.67 & 10.00 & -8.33 & 13.34 & -8.34 & -34.92 \\
        \texttt{Qwen$2.5$-14b} & 3.09 & 7.36 & 4.12 & 6.26 & 4.04 & 3.10 & 2.63 & 0.95 & -11.51 & 7.70 & -5.00 & 4.00 & 4.17 & -8.25 & 11.67 & 5.00 & -10.00 & 13.34 & 1.66 & -50.80 \\
        \texttt{Qwen$2.5$-72b} & 3.09 & 8.42 & 0.00 & 7.30 & 7.07 & 3.10 & 3.51 & 0.00 & -16.82 & 8.79 & 0.00 & 6.00 & 3.12 & -9.28 & 15.00 & 8.33 & -3.33 & 16.67 & 1.66 & -23.81 \\
        \texttt{Qwen$2.5$-7b} & 4.13 & 7.36 & 0.00 & 7.30 & 7.07 & 4.13 & 2.63 & -3.77 & -12.39 & 5.50 & -1.00 & 5.00 & 2.08 & -8.25 & 18.33 & 6.67 & -3.33 & 11.67 & 5.00 & -41.27 \\
        \bottomrule
    \end{tabular}
    }
    \captionsetup[figure]{font=small,labelfont=small}
    \caption{Subset of the average accuracy scores (\%) (higher is better) differential, per phenomenon, of the LLMs against our \texttt{human} baseline. \guillemet{$\Upsilon$} are model that have a specialization in French, while \guillemet{$\Gamma$} are reasoning one.
    }
    \label{tab:resultsdiff}
    \vspace{-1.25em}
\end{table*}

\begin{figure}
    \centering
    \pgfplotstableread{data0 blue
    110655493.0 57.07
    336695557.0 53.89
    110650880.0 61.56
    278295186.0 50.94
    560142482.0 72.69
    13960238080.0 88.02
    13960238080.0 88.02
    8030261248.0 87.05
    671142482 90.01
    671142482 87.34
    671142482 83.02
    6921720704.0 89.32
    124439808.0 58.6
    8030261248.0 89.1
    8030261248.0 88.47
    70553706496.0 88.25
    70553706496.0 87.68
    8030261248.0 87.85
    8030261248.0 88.7
    70553706496.0 89.44
    70553706496.0 88.64
    1235814400.0 85.12
    1235814400.0 82.85
    3212749824.0 88.02
    3212749824.0 86.6
    559214592.0 85.52
    1065314304.0 86.37
    1722408960.0 87.85
    7069016064.0 88.53
    559214592.0 84.1
    1065314304.0 86.37
    494032768.0 83.02
    494032768.0 82.74
    1543714304.0 87.22
    1543714304.0 86.83
    3085938688.0 88.42
    3085938688.0 88.07
    7615616512.0 89.66
    7615616512.0 89.32
    14770033664.0 88.93
    14770033664.0 89.1
    32763876352.0 90.29
    32763876352.0 90.23
    72706203648.0 89.72
    72706203648.0 90.52
    361821120.0 72.35
    361821120.0 70.53
    134515008.0 67.75
    134515008.0 65.59
    1711376384.0 78.25
    1711376384.0 77.74
    3821079552.0 86.6
    3821079552.0 86.31
    2614341888.0 86.26
    2614341888.0 84.95
    9241705984.0 88.36
    9241705984.0 86.77
    27227128320.0 89.1
    27227128320.0 86.54
    8019808256.0 87.96
    7248023552.0 85.86
    7248023552.0 86.54
    22247282688.0 88.7
    12247782400.0 88.25
    46702792704 89.04
    46702792704 88.02
    8028033024.0 78.71
    8028033024.0 77.51
    76961152.0 51.62
    247577856.0 59.8
    783150080.0 60.65
    2849757184.0 63.77
    11135332352.0 62.52
    7615616512.0 76.38
    8030261248.0 83.25
    14770033664.0 87.68
    32763876352.0 87.96
    14659507200.0 89.89
    7615616512.0 88.98
    14770033664.0 88.64
    32763876352.0 89.84
    32763876352.0 89.61
    46702792704.0 89.1
    46702792704.0 89.49
    20905482240.0 86.88
    14765947904.0 88.59
    }\dataZ
    \pgfplotstableread{data1 blue
    76961152.0 64.93714130675127
    110650880.0 66.49488531413537
    110655493.0 66.49506417453468
    124439808.0 66.99875532414447
    134515008.0 67.33277621481867
    134515008.0 67.33277621481867
    247577856.0 69.95011305263532
    278295186.0 70.45190291411052
    336695557.0 71.26920434529441
    361821120.0 71.57798571624579
    361821120.0 71.57798571624579
    494032768.0 72.91422935924865
    494032768.0 72.91422935924865
    559214592.0 73.44594098387797
    559214592.0 73.44594098387797
    560142482.0 73.45305398058177
    783150080.0 74.89089899889905
    1065314304.0 76.21104905572642
    1065314304.0 76.21104905572642
    1235814400.0 76.84799859374093
    1235814400.0 76.84799859374093
    1543714304.0 77.8024393270382
    1543714304.0 77.8024393270382
    1711376384.0 78.24480449976876
    1711376384.0 78.24480449976876
    1722408960.0 78.27237406653119
    2614341888.0 80.06269591748611
    2614341888.0 80.06269591748611
    2849757184.0 80.43261753188855
    3085938688.0 80.77422571119462
    3085938688.0 80.77422571119462
    3212749824.0 80.94700457421507
    3212749824.0 80.94700457421507
    3821079552.0 81.6909793682589
    3821079552.0 81.6909793682589
    6921720704.0 84.24002245949802
    7069016064.0 84.33036435305333
    7248023552.0 84.43765584036667
    7248023552.0 84.43765584036667
    7615616512.0 84.64990952214083
    7615616512.0 84.64990952214083
    7615616512.0 84.64990952214083
    7615616512.0 84.64990952214083
    8019808256.0 84.87177983044836
    8028033024.0 84.87617759258701
    8028033024.0 84.87617759258701
    8030261248.0 84.8773682425427
    8030261248.0 84.8773682425427
    8030261248.0 84.8773682425427
    8030261248.0 84.8773682425427
    8030261248.0 84.8773682425427
    8030261248.0 84.8773682425427
    9241705984.0 85.4802056054519
    9241705984.0 85.4802056054519
    11135332352.0 86.27991604669626
    12247782400.0 86.68845248544132
    13960238080.0 87.24992563568283
    13960238080.0 87.24992563568283
    14659507200.0 87.45962141601726
    14765947904.0 87.49066065794521
    14770033664.0 87.49184764534456
    14770033664.0 87.49184764534456
    14770033664.0 87.49184764534456
    14770033664.0 87.49184764534456
    20905482240.0 88.98236934384144
    22247282688.0 89.24926652765514
    27227128320.0 90.11589521882105
    27227128320.0 90.11589521882105
    32763876352.0 90.91009720322494
    32763876352.0 90.91009720322494
    32763876352.0 90.91009720322494
    32763876352.0 90.91009720322494
    32763876352.0 90.91009720322494
    46702792704.0 92.43093606257118
    46702792704.0 92.43093606257118
    70553706496.0 94.20101475704945
    70553706496.0 94.20101475704945
    70553706496.0 94.20101475704945
    70553706496.0 94.20101475704945
    72706203648.0 94.32995107309426
    72706203648.0 94.32995107309426
    }\dataO
    \resizebox{0.49\textwidth}{!}{%
    \begin{tikzpicture}
    \definecolor{636efa}{HTML}{636efa}
    
    \begin{axis}[
    xmode=log,
    ymin=45,
    ymax=95,
    xmin=66961152.0,
    xmax=12870033664.0,
    xlabel=Model size,
    ylabel=Accuracy (\%),
    xtick pos=left, 
    ytick pos=left, 
    ]
    \addplot+ [mark=*, only marks, mark options={solid, fill=636efa}] table[y=blue] {\dataZ};
    \addplot+ [mark=none, forget plot, blue, line width=0.25mm] table[y=blue] {\dataO};
    \addplot[mark=none, red, dashed, domain=1:12870033664.0, line width=0.25mm] {49.40};
    \addplot[mark=none, Green, dashed, domain=1:12870033664.0, line width=0.25mm] {88.87};
    \end{axis}
    \end{tikzpicture}
    }
    \caption{Comparison between Model size and QFrBLiMP accuracy. 
    The \textcolor{blue}{\textbf{blue solid line}} represents a log-transformed linear data fit, while the \textcolor{Green}{\textbf{green}} and \textcolor{red}{\textbf{red}} dashed represents the human and random baselines.}
    \label{fig:tendancies}
    \vspace{-1.5em}
\end{figure}

\begin{figure}
    \centering
    \pgfplotstableread{data0 y0
    50.98 57.07
    56.86 53.89
    74.12 61.56
    56.08 50.94
    78.04 72.69
    97.65 88.02
    97.65 88.02
    92.94 87.05
    99.22 90.01
    94.51 87.34
    94.51 83.02
    97.65 89.32
    66.67 58.6
    97.65 89.1
    95.69 88.47
    98.43 88.25
    94.12 87.68
    96.47 87.85
    98.04 88.7
    97.25 89.44
    95.69 88.64
    94.9 85.12
    90.98 82.85
    96.86 88.02
    95.69 86.6
    96.86 85.52
    97.65 86.37
    98.43 87.85
    98.43 88.53
    92.94 84.1
    96.08 86.37
    93.33 83.02
    92.94 82.74
    93.73 87.22
    92.94 86.83
    94.9 88.42
    94.9 88.07
    96.47 89.66
    97.25 89.32
    96.86 88.93
    97.65 89.1
    97.65 90.29
    96.47 90.23
    98.82 89.72
    98.43 90.52
    89.02 72.35
    84.71 70.53
    80.78 67.75
    75.69 65.59
    94.51 78.25
    94.12 77.74
    94.9 86.6
    96.86 86.31
    96.86 86.26
    95.69 84.95
    97.25 88.36
    93.73 86.77
    98.04 89.1
    96.47 86.54
    98.43 87.96
    97.65 85.86
    97.25 86.54
    99.61 88.7
    97.25 88.25
    98.04 78.71
    98.04 77.51
    62.75 51.62
    70.59 59.8
    74.51 60.65
    72.16 63.77
    65.88 62.52
    83.92 76.38
    94.9 83.25
    92.94 87.68
    96.08 87.96
    98.04 89.89
    97.65 88.98
    97.65 88.64
    96.86 89.84
    98.04 89.61
    99.61 89.1
    99.61 89.49
    93.73 86.88
    97.25 88.59
    }\dataZ
    \pgfplotstableread{data1 y0
    45 45
    100 100
    }\dataO
    \resizebox{0.49\textwidth}{!}{%
    \begin{tikzpicture}
    
    \definecolor{636efa}{HTML}{636efa}
    
    \begin{axis}[
    xmin=45,
    xmax=100,
    ymin=45,
    ymax=100,
    xlabel=MultiBLiMP,
    ylabel=QFrBLiMP,
    xtick pos=left, 
    ytick pos=left, 
    ]
    \addplot+ [mark=*, only marks, mark options={solid, fill=636efa}, forget plot] table[y=y0] {\dataZ};
    \addplot+ [mark=none] table[y=y0] {\dataO};
    \end{axis}
    \end{tikzpicture}%
    }

    \caption{Comparison of LM performance on the MultiBLiMP and QFrBLiMP benchmarks. The \textcolor{blue}{\textbf{blue solid line}} represents performance parity ($y=x$).}
    \label{fig:versusmulti}
        \vspace{-2em}
\end{figure}

\begin{figure}
    \centering
    \resizebox{0.49\textwidth}{!}{
    \begin{tikzpicture}
    \begin{axis}[
        width=12cm, height=12cm,
        xlabel={MultiBlimp Accuracy (\%)},
        ylabel={QFrBLiMP Accuracy (\%)},
        xmin=0, xmax=100,
        ymin=0, ymax=100,
        xtick pos=left, 
        ytick pos=left, 
        grid=none,
        scatter/classes={
            green_dot={mark=*, draw=green!60!black, fill=green!60!black, opacity=0.5},
            red_dot={mark=*, draw=red!60!black, fill=red!60!black, opacity=0.5},
            blue_dot={mark=*, draw=blue!60!black, fill=blue!60!black, opacity=0.5}
        }
    ]
    
    \addplot [black, dashdotted, domain=0:100] {x};
    
    
    \draw [black!70, dashed, line width=1pt] (axis cs:49.41, 0) -- (axis cs:49.41, 100);
    \draw [black!70, dashed, line width=1pt] (axis cs:0, 49.40) -- (axis cs:100, 49.40);
    
    \addplot [gray, dotted, line width=1.5pt] table [x=x, y=y_upper, col sep=comma] {data/pgf_sig_curves_minimal_pairs.csv};
    \addplot [gray, dotted, line width=1.5pt] table [x=x, y=y_lower, col sep=comma] {data/pgf_sig_curves_minimal_pairs.csv};
    
    \addplot [scatter, only marks, scatter src=explicit symbolic] 
        table [x=acc_multi, y=acc_fr, meta=tex_class, col sep=comma] {data/pgf_scatter_minimal_pairs.csv};
    
    \end{axis}
    \end{tikzpicture}
    
    }
    \caption{Accuracy comparison between QFrBLiMP and MultiBLiMP. The black dash-dotted line represents equal performance. The gray dotted lines represent the statistical significance interval using a Z-test ($\alpha=0.001$). \textcolor{black}{\textbf{Black}} dashed lines are our \texttt{Random} baseline scores. 
    \textcolor{red!60!black}{\textbf{Red}}, \textcolor{green!60!black}{\textbf{green}}, and \textcolor{blue!60!black}{\textbf{blue}} correspond to the model performances from \autoref{tab:resultscomp}.}
    \label{fig:ztest}
\end{figure}

\subsection{Scaling Laws and the Path to Human Performance}
The primary finding of our study is illustrated in \autoref{fig:tendancies}. 
There is a clear and strong positive correlation between model size and grammatical accuracy, confirming that linguistic competence on the QFrBLiMP benchmark is an emergent capability that scales with the logarithm of the model size. 
The log-transformed linear fit (\textcolor{blue}{\textbf{blue solid}}) demonstrates this trend, charting a path from the \texttt{Random} baseline towards the Human performance ceiling (\textcolor{Green}{\textbf{green dashed}}).

The plot also reveals two important nuances. 
First, there is significant performance variance among smaller models, i.e. with fewer than $10^9$ parameters, where architectural choices and training data quality appear to play a larger role. 
Second, as models approach the $10^{10}$ parameter scale, their performance converges more tightly, clustering in the 85-90\% accuracy range. 
It suggests that while scaling is effective, there may be diminishing returns for simply increasing parameter count, with the largest models approaching a performance plateau just below the human baseline.

\subsection{A Hierarchy of Grammatical Competence}
The per-phenomenon results in \autoref{tab:results} break down the aggregate scores of all benchmarked LLMs, revealing a distinct hierarchy of difficulty.

\subsubsection{Mastered Phenomena} 
The core rules of French syntax and morphology are well handled. Phenomena like the \guillemet{-é/-er} distinction (\texttt{12}), \guillemet{clitic placement} (\texttt{5}), and \guillemet{verb flexion} (\texttt{2}) see numerous top-tier models achieving near-perfect scores (95-100\%). 
This indicates these frequent and regular patterns are robustly learned.

For instance, in \autoref{tab:results}, numerous models including \texttt{Bloom-1b7} and \texttt{Lucie-7b $\Upsilon$} score 100.00\% on phenomenon \texttt{12}. 
It not only demonstrates a comprehensive grasp of the rule but also surpasses the human baseline of 93.00\% by a significant margin, as shown in \autoref{tab:resultsdiff}. 
It suggests that for highly regular, frequent, and formally unambiguous grammatical tasks, LLMs' ability to learn and apply patterns exceeds the average consistency of human annotators. 
This pattern of performance suggests that these grammatical structures are well-represented in the pre-training data and have been effectively internalized by LLMs.

\subsubsection{Challenging Phenomena} 
A second tier of phenomena emerges in \autoref{tab:results}, presenting a more substantial challenge. 
This category includes phenomena requiring long-distance dependency tracking and knowledge of complex syntactic configurations.
While the best-performing LMs demonstrate a strong grasp of these rules, performances are more varied and rarely perfect, indicating an incomplete mastery. 
For instance, on \guillemet{past participle agreement} (\texttt{1}), even a top-tier model like \texttt{Qwen2.5-7b} achieves a high but imperfect score of 94.85\%. 
It contrasts with the performance on mastered phenomena, where scores are frequently at or near 100\%.
The challenge here likely stems from the complexity of the underlying rules; indeed, \guillemet{past participle agreement} in French is governed by a non-trivial set of conditions involving the auxiliary verb and the position of the direct object. 
Similarly, \guillemet{syntactic islands} represent abstract constraints on sentence structure that are more complex than simple morphological agreement. 
The models' strong, yet imperfect, performance suggests they have learnt these complex regularities but struggle with the edge cases and intricacies that define full competence.

\subsubsection{Unsolved Phenomena} 
At the top of the difficulty hierarchy lie two phenomena that prove fundamentally challenging for every model tested, regardless of scale, architecture, or specialization: \guillemet{lexical semantics} (\texttt{9}) and \guillemet{orphaned prepositions} (\texttt{20}). 
The LLMs' failure on these tasks is not merely a matter of degree but appears to represent a hard ceiling for current architectures and evaluation methods.

This difficulty is so profound that it seems to defy the scaling laws observed elsewhere. 
As shown in \autoref{fig:tendancies}, increasing model size generally yields better performance, yet even the 72-billion parameter (i.e. \texttt{Qwen2.5-72b}) scores a dismal 60.32\% on \guillemet{orphaned prepositions} (\texttt{20}), barely outperforming the random baseline of 55.56\%. 
The gap to human performance is staggering; \autoref{tab:resultsdiff} reveals that the best models lag the human baseline by more than 20\% on this task.

The source of this failure likely lies in the tasks themselves; indeed, \guillemet{lexical semantics} (\texttt{9}) requires real-world knowledge and an understanding of nuanced word meanings that cannot be resolved by syntactic plausibility alone.
Similarly, \guillemet{orphaned prepositions} (\texttt{20}) test a subtle and abstract syntactic rule that is rare in most training corpora. 
The universal struggle with these phenomena highlights a critical limitation of current models: their difficulty in moving beyond surface-level statistical patterns to grasp deeper semantic and complex syntactic constraints.

\subsection{The Impact of Specialization and Instruction-Tuning}
Our findings suggest that, once scale are accounted for, further specialization in French ($\Upsilon$), a focus on reasoning ($\Gamma$), or instruction tuning (it) does not confer a decisive advantage on this benchmark. These factors appear secondary to the model's foundational properties.

While French-specialized models are highly competitive, they do not establish a distinct, superior tier of performance. 
For example, the specialized 14B parameter \texttt{Chocolatine-2-14b-it} performs on par with the general-domain \texttt{Qwen2.5-14b}, with neither consistently outperforming the other across all phenomena. 
This suggests that the breadth and quality of data in large, general-purpose training runs are sufficient to achieve competitive performance, even in addressing language-specific nuances.

More strikingly, the effect of instruction-tuning (it) appears to be a double-edged sword. 
While it can sometimes lead to minor improvements, it can also be detrimental to formal grammatical knowledge. 
This is best exemplified by the \texttt{Llama-3.1-70b} model pair on \guillemet{Négation standard} (\texttt{7}). 
The base model achieves a 96.49\% accuracy, whereas its instruction-tuned variant, \texttt{Llama-3.1-70b-it}, drops to 85.09\%; a decrease of over 11 percentage points. 
This may suggest that the process of aligning models for conversational abilities can interfere with or overwrite their knowledge of formal grammatical rules, a phenomenon akin to catastrophic forgetting. Or it might be due to a language mismatch between the instruction tuning (in English) and our task (in French). Future research could further investigate this issue.

\subsection{Closing the Gap with Human Judgment}
Analyzing the performance differential against the Human baseline in \autoref{tab:resultsdiff} reveals a divergence between the grammatical competence of LLMs and that of humans. 

On the one hand, models exhibit a strong, consistent mastery of tasks involving abstract formal rules. This is most evident with syntactic island constraints, particularly \guillemet{subject islands} (\texttt{15}) and \guillemet{adjunct islands} (\texttt{16}), as well as frequent, pattern-based rules like the \guillemet{-é/-er} distinction (\texttt{12}). On these tasks, many top-tier models, such as \texttt{Gemma-2-9b}, outperform the human baseline by 20\% or more. 
It suggests that once models internalize a statistical or abstract rule from vast data, they can apply it with a logical precision that surpasses the human baseline, who may be influenced by semantic plausibility or other heuristics.

On the other hand, for phenomena that depend on deep semantic or pragmatic understanding, the gap is inverse. 
The most challenging tasks for models, and the ones where they lag furthest behind human performance, are \guillemet{lexical semantics} (\texttt{9}) and \guillemet{orphaned prepositions} (\texttt{20}).
Even the largest models exhibit differentials of around -10\% for lexical semantics, whereas for orphaned prepositions, the deficit ranges from -20\% to -30\%. 
It indicates that while models excel at formal rule application, they still lack the rich, context-aware reasoning that humans use to resolve semantic ambiguities and complex syntactic structures.
This dramatic failure highlights a reliance on surface-level statistical regularities rather than on the abstract, context-aware reasoning that humans employ to handle semantic ambiguities and complex syntactic structures. While LLMs are powerful at internalizing frequent co-occurrence patterns, they struggle to acquire the more profound grammatical knowledge needed for complex cases. This distinction marks a critical frontier in bridging the gap between artificial and human LC.

\subsection{QFrBLiMP Versus MultiBLiMP}
An analysis of \autoref{fig:versusmulti} reveals two primary insights into the relationship between our QFrBLiMP benchmark and MultiBLiMP. 
First, the strong positive correlation between the scores confirms that both benchmarks measure related aspects of grammatical competence, validating the relevance of QFrBLiMP. 
More critically, however, the plot shows a systematic downward shift in performance, with nearly all models scoring lower on QFrBLiMP than on MultiBLiMP, as evidenced by the cluster of points below the parity line ($y=x$). 
We hypothesize that this indicates QFrBLiMP is a more challenging benchmark. By carefully controlling for lexical artifacts and ensuring each MP isolates a single linguistic phenomenon, QFrBLiMP likely reduces the availability of superficial statistical cues that models might exploit in MultiBLiMP. Furthermore, the significantly larger size of our dataset (1,761 versus 255) implies a broader coverage of linguistic phenomena, potentially exposing more syntactic weaknesses in the models.

Moreover, the statistical analysis presented in \autoref{fig:ztest} provides deeper insight into this performance gap. 
The gray dotted lines delineate the significance interval derived from the Z-test ($\alpha = 0.001$); points falling within this band represent models for which the performance difference between the two dialects is not statistically distinguishable. 
Notably, the models colored in \textbf{blue} fall well below this interval, confirming a statistically significant performance degradation on QFrBLiMP compared to MultiBLiMP. 
It indicates that, for this group, lower scores on Quebec French are not due to random chance but reflect a genuine lack of robustness in handling dialectal variation. 
In contrast, the models in \textbf{green}, which are clustered in the upper-right quadrant of high accuracy, largely remain within or near the significance band. 
It suggests that the most capable models demonstrate better cross-dialectal generalization, maintaining comparable grammatical competence across both Metropolitan and Quebec French varieties.

\section{Conclusion and Future Works}
\label{sec:conclusion}

In this paper, we introduce QFrBLiMP, the first LMP benchmark specifically designed for Quebec French. 
It includes 1,761 MPs across 20 distinct grammatical phenomena from the BDL, an official, human-made normative linguistic resource.
It also includes grammatical judgments from twelve human annotators.
This resource is a novel and valuable tool for evaluating LLMs' competency in Quebec-French grammar. 
Our grammar-driven approach ensures that the benchmark tests for knowledge of attested linguistic rules specific to this variety of French.

Moreover, we benchmarked \llm{} open-source LLMs. 
Our experimentation yielded several key insights into the nature of grammatical competence in these models. 
We confirmed a strong positive correlation between model scale and overall accuracy, though with apparent diminishing returns. 
Crucially, our analysis revealed a clear hierarchy of difficulty: while models have mastered frequent, regular syntactic and morphological rules, they are challenged by more complex configurations, such as long-distance dependencies, and consistently fail on phenomena requiring deep semantic understanding, including lexical semantics and orphaned prepositions. 
Furthermore, we demonstrated the paramount importance of modern architectures. We showed that language specialization or instruction-tuning does not guarantee superior performance on formal grammar, and that the latter can sometimes be detrimental.

Finally, our comparative analysis with MultiBLiMP highlights a general shift in difficulty in QFrBLiMP. 
Statistical testing confirms that while many models suffer a significant performance degradation when processing the Quebec dialect, the most advanced models demonstrate robust cross-dialectal generalization, falling within the statistical equivalence zone for both varieties.

Our future work could address the lack of effect of language specialization by developing a training corpus to optimize LLM on Quebec-French. 
Moreover, we plan to investigate alternative evaluation methods, such as targeted probing or semantic similarity scores, for phenomena that yield low accuracies.
Finally, we plan to investigate the mechanisms causing the degradation of formal grammatical knowledge during instruction-tuning, a phenomenon observed in our results. We then aim to develop novel alignment techniques that can enhance conversational abilities without sacrificing the model's foundational LC.

\section*{Limitations}
The methodology of QFrBLiMP involves manually extracting sentence pairs from the BDL, an official grammatical resource from the OQLF. 
While this ensures high-quality, linguistically vetted examples, it also introduces a potential distributional shift between the evaluation data and the vast, descriptive corpora typically used to pre-train LLMs.
As the creators of RuBLiMP \citep{taktasheva2024rublimp} note, domain shifts can introduce bias, and the language in QFrBLiMP may not be representative of naturally occurring text. 
The BDL's purpose is pedagogical: to provide clear, often formal, examples that unambiguously illustrate a specific grammatical rule. 
This didactic style may differ significantly from the more varied and informal language found in web-scale pre-training datasets. 

A significant consideration for the QFrBLiMP benchmark is the potential for data leakage, given that its sole source, the BDL, is a public online resource. 
Since LLMs are trained on extensive web data, there is a risk that sentences from the BDL were included in the models' pre-training corpora.
This issue of \guillemet{test data contamination} \citep{jacovi2023stop} can compromise the validity of the benchmark, as a model might achieve high accuracy by recalling sentences from its training data rather than by demonstrating genuine grammatical understanding.

A limitation of the QFrBLiMP benchmark is the inclusion of \guillemet{Anglicism} as a core evaluation category, which differs from the formal grammatical phenomena typically found in such benchmarks \citep{warstadt2020blimp, someya2023jblimp, suijkerbuijk2024blimp, taktasheva2024rublimp, jumelet2025multiblimp}. 
The acceptability of \guillemet{Anglicisms} often pertains to prescriptive stylistics and lexical choice rather than fundamental grammatical correctness. 
This prescriptive tendency is not a recent development but is deeply rooted in the history of the French language. Since the 17th century, institutions like the Académie française have worked to standardize the language, often by excluding regional or popular vocabulary in favour of a more formal, courtly standard, meaning French has not evolved as freely or naturally as other languages \citep{walter2016norme}. 
Unlike violations of syntax (e.g. subject-verb agreement) or morphology, the preference for a native French term over an English loanword can be subjective and dependent on register, context, and evolving language norms. This situation is due to Quebec's historical language and identity fight of the Quebec nation \citep{dickinson2008short}.

\section*{Ethical Considerations}
Another ethical consideration is the potential for inherent biases within the source data. 
QFrBLiMP sentences are derived from the BDL, a normative and official linguistic resource for Quebec-French. 
While this ensures grammatical correctness according to a specific standard, it may also embed institutional or prescriptive biases.
Moreover, the OQLF may be considered a political institution, given that Quebec is the only French-speaking nation in North America; thus, some MPs might not be accepted elsewhere as valid French \citep{hambye2014language}.

As with any tool for evaluating and improving LLM, the potential for misuse must be considered. 
Indeed, improving text generation quality through such benchmarks could lead to the misuse of LLMs for harmful purposes, such as disinformation, harmful text generation, and online harassment \citep{vessey2016language, mcdougall2019connecting, weidinger2021ethical, bender2021dangers}.
While the intended use of QFrBLiMP is for research and development, the creators of such resources bear a responsibility to acknowledge that improvements spurred by their work could be used in both research and development.

\section*{Acknowledgements}
This research was made possible thanks to the support of a Canadian insurance company, NSERC research grant RDCPJ 537198-18 and FRQNT doctoral research grant. We thank the reviewers for their comments regarding our work.

\bibliography{anthology,custom}
\bibliographystyle{acl_natbib}

\appendix
\section{Linguistic Phenomena Examples}
\label{ann:phenomena}

We present in \autoref{tab:categorization_complete_Ex} an example, its \underline{error}, and translations for each linguistic phenomenon. 
We also present in \autoref{ann:description} the translated description of our LP.

\begin{table*}
    \centering
    \tiny
    \begin{tabular}{cp{0.4\textwidth}cp{0.4\textwidth}}
        \toprule
        LP &  Example & LP & Example\\
        \midrule
        \texttt{1}      
        &\begin{tabular}[c]{@{}p{0.4\textwidth}@{}}
        \textit{Après tous ces efforts, elles se sont senties très fatiguées.}\\
        \textit{Après tous ces efforts, elles se sont \underline{senti} très fatiguées.}\\
        After all that effort, they felt very tired.\\
        After all that effort, they \underline{felt} very tired.
        \end{tabular} &
        \texttt{11}
        &\begin{tabular}[c]{@{}p{0.4\textwidth}@{}}
        \textit{À cette époque, c’était considéré comme des mœurs dissolues.}\\
        \textit{À cette époque, c’était considéré comme des mœurs \underline{dissolue}.}\\
        In those days, it was considered dissolute.\\
        In those days, it was considered \underline{dissolute}.
        \end{tabular}\\
        \texttt{2}        
        &\begin{tabular}[c]{@{}p{0.4\textwidth}@{}}
        \textit{Ainsi soit-il.}\\
        \textit{Ainsi \underline{soient}-il.}\\
        So be it.\\
        So \underline{be} it.
        \end{tabular}&
        \texttt{12}
        &\begin{tabular}[c]{@{}p{0.4\textwidth}@{}}
        \textit{À quelle heure avez-vous mangé?}\\
        \textit{À quelle heure avez-vous \underline{manger}?}\\
        What time did you eat?\\
        What time did you \underline{eat}?
        \end{tabular}\\
        \texttt{3}     
        &\begin{tabular}[c]{@{}p{0.4\textwidth}@{}}
        \textit{À l’avenir, nous ne soulignerons que l’anniversaire des membres du personnel qui le souhaitent.}\\
        \textit{À l’avenir, nous \underline{ne soulignerons l’anniversaire} des membres que du personnel qui le souhaitent.}\\
        In future, we will only celebrate the birthdays of staff members who wish to do so.\\
        In future, we will \underline{only celebrate the birthdays} of staff members who wish to do so.
        \end{tabular} &
        \texttt{13}
        &\begin{tabular}[c]{@{}p{0.4\textwidth}@{}}
        \textit{\guillemettwo{Merci de m’avoir aidé à déménager. — De rien!}}\\
        \textit{\guillemettwo{Merci \underline{pour m’avoir} aidé à déménager. — De rien!}}\\
        \guillemet{Thank you for helping me move. - You're welcome!}\\
        \guillemet{Thank you \underline{for helping} me move. - You're welcome}
        \end{tabular}\\
        \texttt{4}        
        &\begin{tabular}[c]{@{}p{0.4\textwidth}@{}}
        \textit{Autant en finir maintenant que de remettre cela à plus tard.}\\
        \textit{Autant en finir maintenant \underline{que remettre} cela à plus tard.}\\
        It's better to get it over with now than to put it off.\\
        It's better to get it over with now than \underline{to put it} off.
        \end{tabular}&
        \texttt{14}
        &\begin{tabular}[c]{@{}p{0.4\textwidth}@{}}
        \textit{Cette regrettable situation devait ne pas arriver.}\\
        \textit{Cette regrettable situation devait \underline{n'arriver pas}.}\\
        This unfortunate situation was never meant to happen.\\
        This unfortunate situation \underline{was meant never} to happen.
        \end{tabular}\\
        \texttt{5}     
        &\begin{tabular}[c]{@{}p{0.4\textwidth}@{}}
        \textit{Bien qu'en proie à une forme d'insécurité linguistique, certains locuteurs s’efforcent de ne pas le laisser paraître.}\\
        \textit{Bien qu'en proie à une forme d'insécurité linguistique, certains locuteurs s’efforcent \underline{de ne le pas} laisser paraître.}\\
        Although plagued by a form of linguistic insecurity, some speakers try not to let it show.\\
        Although plagued by a form of linguistic insecurity, some speakers try \underline{not to let it} show.
        \end{tabular}&
        \texttt{15}
        &\begin{tabular}[c]{@{}p{0.4\textwidth}@{}}
        \textit{La mélodie que l’oreille qui la reçoit apprécie touche l'âme.}\\
        \textit{La mélodie que l’oreille \underline{qui reçoit} apprécie touche l'âme.}\\
        The melody appreciated by the ear touches the soul.\\
        The melody \underline{that the receiving} ear appreciates touches the soul.
        \end{tabular}\\
        \texttt{6}       
        &\begin{tabular}[c]{@{}p{0.4\textwidth}@{}}
        \textit{Ces documents, je ne veux certainement plus les voir.}\\
        \textit{Ces documents, je \underline{ne les veux certainement plus voir}.}\\
        I certainly don't want to see these documents again.\\
        I certainly don't want to \underline{see these documents again}.
        \end{tabular}&
        \texttt{16}
        &\begin{tabular}[c]{@{}p{0.4\textwidth}@{}}
        \textit{Martin a mangé le gâteau qui cuisait pendant que Noella le tournait.}\\
        \textit{Martin a mangé le gâteau qui cuisait pendant que Noella\underline{ tournait}.}\\
        Martin ate the baking cake while Noella turned it.\\
        Martin ate the baking cake while Noella was\underline{ turned}.
        \end{tabular}\\
        \texttt{7}       
        &\begin{tabular}[c]{@{}p{0.4\textwidth}@{}}
        \textit{\guillemettwo{Mais madame, vous m’aviez dit que je ne pouvais pas compter sur vous!}}\\
        \textit{\guillemettwo{Mais madame, vous m’aviez dit que \underline{je pouvais pas} compter sur vous!}}\\
        \guillemet{But Madame, you told me I couldn't count on you!}\\
        \guillemet{But Madame, you told me \underline{I couldn't} count on you}\\
        \end{tabular}&
        \texttt{17}
        &\begin{tabular}[c]{@{}p{0.4\textwidth}@{}}
        \textit{Précisez-nous ce que ignorez si vous le voulez.}\\
        \textit{Précisez-nous ce que ignorez si \underline{vous voulez}.}\\
        Tell us what you don't know if you like.\\
        Tell us what you don't know if \underline{you the like}.
        \end{tabular}\\
        \texttt{8}
        &\begin{tabular}[c]{@{}p{0.4\textwidth}@{}}
        \textit{À compter de l’année prochaine, \underline{de} nouveaux règlements municipaux devront être respectés.}\\
        \textit{À compter de l’année prochaine, \underline{des} nouveaux règlements municipaux devront être respectés.}\\
        Starting next year, new municipal by-laws will have to be respected.\\
        Starting next year,\underline{ } new municipal by-laws will have to be respected.
        \end{tabular}&
        \texttt{18}
        &\begin{tabular}[c]{@{}p{0.4\textwidth}@{}}
        \textit{À l’époque, le mouvement surréaliste que le refus que la convention n'étouffe nourissait en était à ses balbutiements.}\\
        \textit{À l’époque, le mouvement surréaliste que le refus que la convention \underline{ne l'}étouffe nourissait en était à ses balbutiements.}\\
        At the time, the Surrealist movement, fueled by a refusal to be stifled by convention, was in its infancy.\\
        At the time, the Surrealist movement, nurtured by the refusal of convention \underline{to stifle it}, was in its infancy.
        \end{tabular}\\
        \texttt{9}
        &\begin{tabular}[c]{@{}p{0.4\textwidth}@{}}
        \textit{Ancien toxicomane, il a cessé toute consommation de stupéfiants.}\\
        \textit{Ancien \underline{addict}, il a cessé toute consommation de stupéfiants.}\\
        A former drug addict, he no longer uses drugs.\\
        A former \underline{addict}, he no longer uses drugs.
        \end{tabular}&
        \texttt{19}
        &\begin{tabular}[c]{@{}p{0.4\textwidth}@{}}
        \textit{Ce point, dont les inconvénients nuisent à la mise en valeur de ses avantages, sera discuté lors de la prochaine réunion.}\\
        \textit{Ce point, dont les inconvénients nuisent à la mise en valeur \underline{des} avantages, sera discuté lors de la prochaine réunion.}\\
        This point, whose disadvantages detract from its advantages, will be discussed at the next meeting.\\
        This point, whose disadvantages detract \underline{from the} advantages, will be discussed at the next meeting.
        \end{tabular}\\
        \texttt{10}
        &\begin{tabular}[c]{@{}p{0.4\textwidth}@{}}
        \textit{À chacun son dû.}\\
        \textit{À chacun \underline{leur} dû.}\\
        To each his own.\\
        To each \underline{his} own.
        \end{tabular}&
        \texttt{20}
        &\begin{tabular}[c]{@{}p{0.4\textwidth}@{}}
        \textit{La poutre que Michel s'est cogné la tête contre semble très dure.}\\
        \textit{La poutre que Michel s'est cogné la tête contre \underline{elle} semble très dure.}\\
        The beam Michel hit his head on seems very hard.\\
        The beam Michel hit his head on \underline{she} seems very hard.
        \end{tabular}
        \\\bottomrule    
        \end{tabular}%
    \caption{Example of pair, per linguistic phenomena (\guillemet{LP}) with the error \underline{underlined}.}
    \label{tab:categorization_complete_Ex}
\end{table*}

\begin{table*}
    \centering
    \tiny
    \begin{tabular}{cp{0.4\textwidth}cp{0.4\textwidth}}
        \toprule
        LP &  Description & LP & Description\\
        \midrule
        \texttt{1} & The agreement in gender and number of the past participle based on the auxiliary verb used and the position of the direct object. &
        \texttt{11} & The agreement of an adjective in gender and number with the noun it modifies.
        \\
        \texttt{2} & The agreement of the verb in number and person with its subject, particularly in the subjunctive mood. &
        \texttt{12} & The correct use of the past participle ending (e.g. -é) versus the infinitive ending (e.g. -er), especially in compound tenses.
        \\
        \texttt{3} & The correct placement of the restrictive expression \textit{ne... que}, which must frame the element being restricted. &
        \texttt{13} & The correct selection of the preposition (e.g. \textit{de} or \textit{pour}) that is required by a specific verb or expression to introduce its complement.
        \\
        \texttt{4} & The required use of a preposition, such as \textit{de}, to correctly link parts of a correlative construction, especially before an infinitive. &
        \texttt{14} & The rule for negating an infinitive, where the negative particles (\textit{ne pas}) must be placed together directly before the infinitive verb.
        \\
        \texttt{5} & The correct placement of a clitic pronoun in relation to negative particles when modifying an infinitive verb. &
        \texttt{15} & A syntactic constraint (Subject Island) that forbids extracting an element from a relative clause that modifies the subject of a sentence.
        \\
        \texttt{6} & A rule where a clitic pronoun that is the object of an infinitive moves to a position before the main, governing verb. &
        \texttt{16} & A syntactic constraint (Adjunct Island) that forbids extracting an element from an adverbial or adjunct clause.
        \\
        \texttt{7} & The formal requirement of using the two-part negation (e.g. \textit{ne... pas}), where omitting the \textit{ne} is considered non-standard. &
        \texttt{17} & A syntactic constraint (Wh-Island) that forbids extracting an element from an embedded interrogative clause.
        \\
        \texttt{8} & The rule where the plural indefinite determiner \textit{des} changes to \textit{de} when placed before an adjective that precedes the noun. &
        \texttt{18} & A syntactic constraint (complex noun phrase island) that forbids extracting an element from a relative clause that modifies a noun.
        \\
        \texttt{9} & The use of a standard French term as prescribed by normative lexical standards over an anglicism or other non-standard term. &
        \texttt{19} & The correct use of the relative pronoun \textit{dont} to replace a complement of a noun introduced by the preposition \textit{de}.
        \\
        \texttt{10} & The rules of agreement for pronouns and possessives within fixed or idiomatic expressions. &
        \texttt{20} & The rule against preposition stranding, requiring a preposition to be followed by its object, sometimes using a resumptive pronoun.
        \\\bottomrule    
    \end{tabular}%
    \caption{Translated linguistic phenomena (\guillemet{LP}).}
    \label{ann:description}
\end{table*}

\section{Annotation Interface}
\label{ann:clf_annotation}

\begin{figure*}
    \centering
    \includegraphics[width=\textwidth]{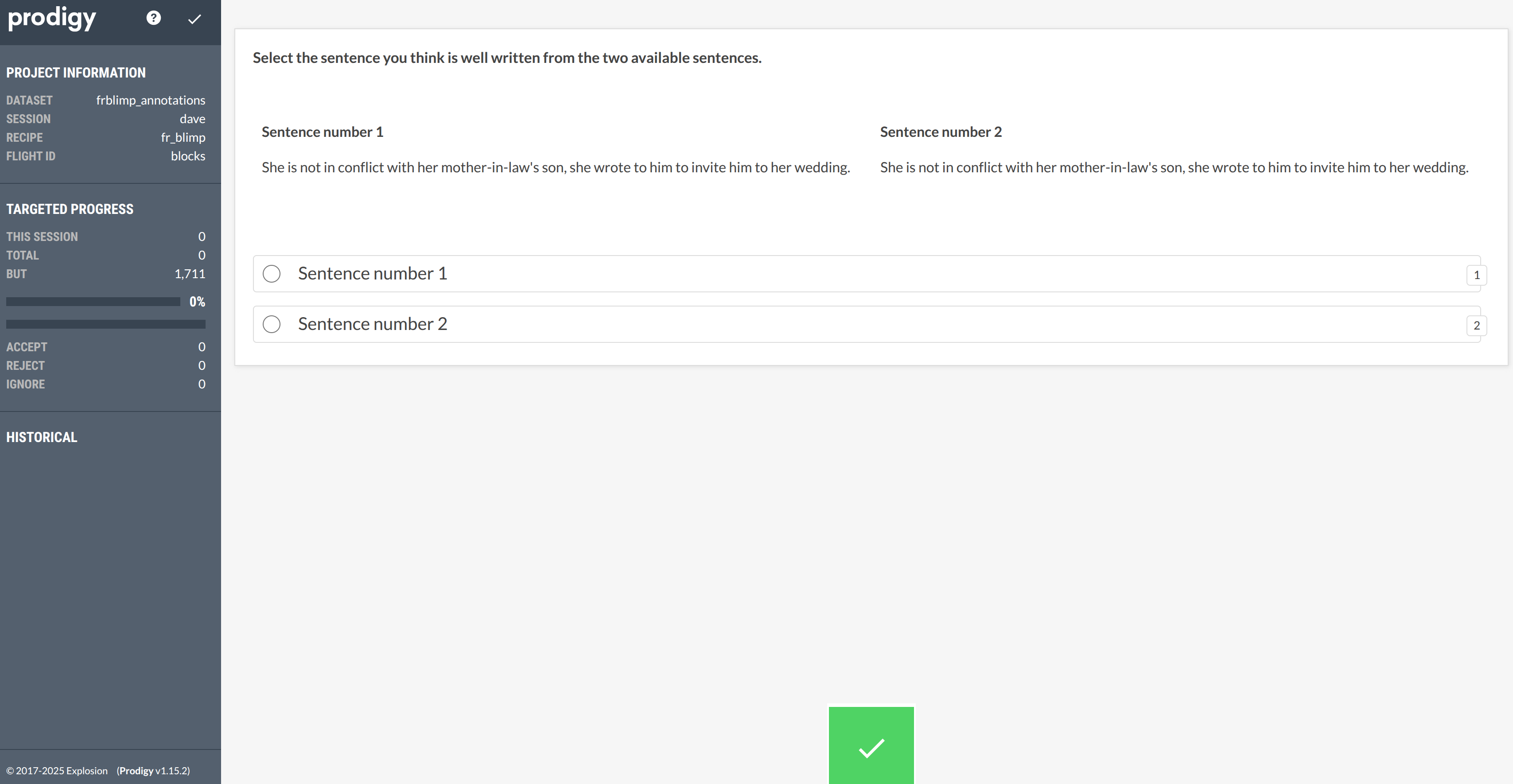}
    \caption{The Prodigy annotation interface (in French) used by the annotators to evaluate the MPs.}
    \label{fig:clf_annotation}
\end{figure*}

The \autoref{fig:clf_annotation} presents the evaluation interface used by our annotators (in French). It is a custom adaptation of the Prodigy annotation tool \citep{prodigy_montani_honnibal}. 

\clearpage
\section{Human Evaluation Datasheet}
\label{ann:hed}

\subsection{Paper and Supplementary Resources (Questions 1.1--1.3)}\label{sec:paper-resources}

\subsubsection*{\qsecbox{Question 1.1: Link to paper reporting the evaluation experiment. If the paper reports more than one experiment, state which experiment you're completing this sheet for. Or, if applicable, enter `for preregistration.'}}
\noindent For preregistration.

\subsubsection*{\qsecbox{Question 1.2: Link to website providing resources used in the evaluation experiment (e.g.\ system outputs, evaluation tools, etc.). If there isn't one, enter `N/A'.}}
\noindent N/A.

\subsubsection*{\qsecbox{Question 1.3: Name, affiliation and email address of person completing this sheet, and of contact author if different.}}
\noindent \textbf{Retracted for double-anonymized anonymity.}

\subsection{System (Questions 2.1--2.5)}\label{sec:system}
\subsubsection*{\qsecbox{Question 2.1: What type of input do the evaluated system(s) take? Select all that apply. If none match, select `Other' and describe.}}\label{sec:input}

\noindent\textit{Check-box options (select all that apply)}: 
\begin{enumerate}[itemsep=0cm, leftmargin=0.5cm, label={\small $\square$}]
    \item[\checkmark] \egcvalue{raw/structured data}: numerical, symbolic, and other data, possibly structured into trees, graphs, graphical models, etc. May be the input e.g.\ to Referring Expression Generation (REG), end-to-end text generation, etc. {NB}: excludes linguistic structures.
    \item \egcvalue{deep linguistic representation (DLR)}: any of a variety of deep, underspecified, semantic representations, such as abstract meaning representations \citep[AMRs;][]{banarescu-etal-2013-abstract} or discourse representation structures \citep[DRSs;][]{kamp2013discourse}.
    \item \egcvalue{shallow linguistic representation (SLR)}: any of a variety of shallow, syntactic representations, e.g.\ Universal Dependency (UD) structures; typically the input to surface realisation.
    \item \egcvalue{text: subsentential unit of text}: a unit of text shorter than a sentence, e.g.\ Referring Expressions (REs), verb phrase, text fragment of any length; includes titles/headlines.
    \item[\checkmark] \egcvalue{text: sentence}: a single sentence (or set of sentences).
    \item \egcvalue{text: multiple sentences}: a sequence of multiple sentences, without any document structure (or a set of such sequences). 
    \item \egcvalue{text: document}: a text with document structure, such as a title, paragraph breaks or sections, e.g.\ a set of news reports for summarisation.
    \item \egcvalue{text: dialogue}: a dialogue of any length, excluding a single turn which would come under one of the other text types.
    \item \egcvalue{text: other}: input is text but doesn't match any of the above \textit{text:*} categories.
    \item \egcvalue{speech}: a recording of speech.
    \item \egcvalue{visual}: an image or video.
    \item \egcvalue{multi-modal}: catch-all value for any combination of data and/or linguistic representation and/or visual data etc.
    \item \egcvalue{control feature}: a feature or parameter specifically present to control a property of the output text, e.g.\ positive stance, formality, author style.
    \item \egcvalue{no input (human generation)}: human generation\footnote{\label{human-generation}We use the term `human generation' where the items being evaluated have been created manually, rather than generated by an automatic system.}, therefore no system inputs.
    \item \egcvalue{other (please specify)}: if input is none of the above, choose this option and describe it.   
\end{enumerate}

\subsubsection*{\qsecbox{Question 2.2: What type of output do the evaluated system(s) generate? Select all that apply. If none match, select `Other' and describe.}}\label{sec:output}

\noindent\textit{Check-box options (select all that apply)}: 

\begin{enumerate}[itemsep=0cm,leftmargin=0.5cm,label={\small $\square$}]
            \item[\checkmark] \egcvalue{raw/structured data}: numerical, symbolic, and other data, possibly structured into trees, graphs, graphical models, etc. May be the input e.g.\ to Referring Expression Generation (REG), end-to-end text generation, etc. {NB}: excludes linguistic structures.
            
            \item \egcvalue{deep linguistic representation (DLR)}: any of a variety of deep, underspecified, semantic representations, such as abstract meaning representations \citep[AMRs;][]{banarescu-etal-2013-abstract} or discourse representation structures \citep[DRSs;][]{kamp2013discourse}.
            
            \item \egcvalue{shallow linguistic representation (SLR)}: any of a variety of shallow, syntactic representations, e.g.\ Universal Dependency (UD) structures; typically the input to surface realisation.
            
            \item \egcvalue{text: subsentential unit of text}: a unit of text shorter than a sentence, e.g.\ Referring Expressions (REs), verb phrase, text fragment of any length; includes titles/headlines.
            
            \item \egcvalue{text: sentence}: a single sentence (or set of sentences).
            
            \item \egcvalue{text: multiple sentences}: a sequence of multiple sentences, without any document structure (or a set of such sequences). 
            
            \item \egcvalue{text: document}: a text with document structure, such as a title, paragraph breaks or sections, e.g.\ a set of news reports for summarisation.
            
            \item \egcvalue{text: dialogue}: a dialogue of any length, excluding a single turn which would come under one of the other text types.
            
           \item \egcvalue{text: other}: select if output is text but doesn't match any of the above \textit{text:*} categories.
            
            \item \egcvalue{speech}: a recording of speech.
            
            \item \egcvalue{visual}: an image or video.
            
            \item \egcvalue{multi-modal}: catch-all value for any combination of data and/or linguistic representation and/or visual data etc.
            
            \item \egcvalue{human-generated `outputs'}: manually created stand-ins exemplifying outputs.
            
            \item \egcvalue{other (please specify)}: if output is none of the above, choose this option and describe it. 
            
        \end{enumerate}

\subsubsection*{\qsecbox{Question 2.3: How would you describe the task that the evaluated system(s) perform in mapping the inputs in Q2.1 to the outputs in Q2.2? Occasionally, more than one of the options below may apply. If none match, select `Other' and describe.}}\label{sec:task}

\noindent\textit{Check-box options (select all that apply)}:  

\begin{enumerate}[itemsep=0cm,leftmargin=0.5cm,label={\small $\square$}]
    \item[\checkmark] \egcvalue{content selection/determination}: selecting  the specific content that will be expressed in the generated text from a representation of possible content. This could be attribute selection for REG (without the surface realisation step). Note that the output here is not text.
    
    \item \egcvalue{content ordering/structuring}: assigning an order and/or structure to content to be included in generated text. Note that the output here is not text.
    
    \item \egcvalue{aggregation}: converting inputs (typically \textit{deep linguistic representations} or \textit{shallow linguistic representations}) in some way in order to reduce redundancy (e.g.\  representations for `they like swimming', `they like running' $\rightarrow$ representation for `they like swimming and running').
   
    \item \egcvalue{referring expression generation}: generating \textit{text} to refer to a given referent, typically represented in the input as a set of attributes or a linguistic representation. 
    
    \item \egcvalue{lexicalisation}: associating (parts of) an input representation with specific lexical items to be used in their realisation. 
    
    \item \egcvalue{deep generation}: one-step text generation from \textit{raw/structured data} or \textit{deep linguistic representations}. One-step means that no intermediate representations are passed from one independently run module to another.
    
    \item \egcvalue{surface realisation (SLR to text)}: one-step text generation from \textit{shallow linguistic representations}. One-step means that no intermediate representations are passed from one independently run module to another.
    
    \item \egcvalue{feature-controlled text generation}: generation of text that varies along specific dimensions where the variation is controlled via \textit{control feature}s specified as part of the input. Input is a non-textual representation (for feature-controlled text-to-text generation select the matching text-to-text task). 
    
    \item \egcvalue{data-to-text generation}: generation from \textit{raw/structured data} which may or may not include some amount of content selection as part of the generation process. Output is likely to be \textit{text:*} or \textit{multi-modal}.
    
    \item \egcvalue{dialogue turn generation}: generating a dialogue turn (can be a greeting or closing) from a representation of dialogue state and/or last turn(s), etc. 

    \item \egcvalue{question generation}: generation of questions from given input text and/or knowledge base such that the question can be answered from the input.
    
    \item \egcvalue{question answering}: input is a question plus optionally a set of reference texts and/or knowledge base, and the output is the answer to the question.
    
    \item \egcvalue{paraphrasing/lossless simplification}: text-to-text generation where the aim is to preserve the meaning of the input while changing its wording. This can include the aim of changing the text on a given dimension, e.g.\ making it simpler, changing its stance or sentiment, etc., which may be controllable via input features. Note that this task type includes meaning-preserving text simplification (non-meaning preserving simplification comes under \textit{compression/lossy simplification} below).
    
    \item \egcvalue{compression/lossy simplification}: text-to-text generation that has the aim to generate a shorter, or shorter and simpler, version of the input text. This will normally affect meaning to some extent, but as a side effect, rather than the primary aim, as is the case in \textit{summarisation}.
    
    \item \egcvalue{machine translation}: translating text in a source language to text in a target language while maximally preserving the meaning. 
    
    \item \egcvalue{summarisation (text-to-text)}: output is an extractive or abstractive summary of the important/relevant/salient content of the input  document(s).

    \item \egcvalue{end-to-end text generation}: use this option if the single system task corresponds to more than one of tasks above, implemented either as separate modules pipelined together, or as one-step generation, other than \textit{deep generation} and \textit{surface realisation}.
    
    \item \egcvalue{image/video description}: input includes \textit{visual}, and the output describes it in some way.
    
    \item \egcvalue{post-editing/correction}: system edits and/or corrects the input text (typically itself the textual output from another system) to yield an improved version of the text.
       
    \item \egcvalue{other (please specify)}: if task is none of the above, choose this option and describe it.
    \end{enumerate}

\subsubsection*{\qsecbox{Question 2.4: Input Language(s), or `N/A'.}}
Quebec-French.
        
\subsubsection*{\qsecbox{Question 2.5: Output Language(s), or `N/A'.}}
N/A

\subsection{Output Sample, Evaluators, Experimental Design}\label{sec:design}

\subsubsection{Sample of system outputs (or human-authored stand-ins) evaluated (Questions 3.1.1--3.1.3)}

\subsubsection*{\qsecbox{Question 3.1.1: How many system outputs (or other evaluation items) are evaluated per system in the evaluation experiment? Answer should be an integer.}}
1761.

\vspace{-.3cm}
\subsubsection*{\qsecbox{Question 3.1.2: How are system outputs (or other evaluation items) selected for inclusion in the evaluation experiment? If none match, select `Other' and describe.}}
\noindent\textit{Multiple-choice options (select one)}:  
\begin{enumerate}[itemsep=0cm,leftmargin=0.5cm,label={\LARGE $\circ$}]
    \item \egcvalue{by an automatic random process from a larger set}: outputs were selected for inclusion in the experiment by a script using a pseudo-random number generator; don't use this option if the script selects every $n$th output (which is not random). 
    \item \egcvalue{by an automatic random process but using stratified sampling over given properties}: use this option if selection was by a random script as above, but with added constraints ensuring that the sample is representative of the set of outputs it was selected from, in terms of given properties, such as sentence length, positive/negative stance, etc.
    \item \egcvalue{by manual, arbitrary selection}: output sample was selected by hand, or automatically from a manually compiled list, without a specific selection criterion.
    \item[\checkmark] \egcvalue{by manual selection aimed at achieving balance or variety relative to given properties}: selection by hand as above, but with specific selection criteria, e.g.\ same number of outputs from each time period.
    \item \egcvalue{Other (please specify)}: if selection method is none of the above, choose this option and describe it.
\end{enumerate}

\subsubsection*{\qsecbox{Question 3.1.3: What is the statistical power of the sample size?}}

\noindent Following the methodology of \citet{card-etal-2020-little}, we obtained a statistical power of X on the output sample w.r.t the automatic evaluation metrics, the two best-performing models (X and Y). We used their online script to estimate the statistical power.

\subsubsection{Evaluators (Questions 3.2.1--3.2.4)}

\subsubsection*{\qsecbox{Question 3.2.1:  How many evaluators are there in this experiment? Answer should be an integer.}}

Twelve.

\subsubsection*{\qsecbox{Question 3.2.2:  What kind of evaluators are in this experiment? Select all that apply. If none match, select `Other' and describe. In all cases, provide details in the text box under `Other'.}}

\noindent\textit{Check-box options (select all that apply)}:  

\begin{enumerate}[itemsep=0cm,leftmargin=0.5cm,label={\small $\square$}]
    \item \egcvalue{experts}: participants are considered domain experts, e.g.\ meteorologists evaluating a weather forecast generator, or nurses evaluating an ICU report generator.
    \item[\checkmark] \egcvalue{non-experts}: participants are not domain experts.
    \item[\checkmark] \egcvalue{paid (including non-monetary compensation such as course credits)}: participants were given some form of compensation for their participation, including vouchers, course credits, and reimbursement for travel unless based on receipts.
    \item \egcvalue{not paid}: participants were not given compensation of any kind.
    \item \egcvalue{previously known to authors}: (one of the) researchers running the experiment knew some or all of the participants before recruiting them for the experiment.
    \item[\checkmark] \egcvalue{not previously known to authors}: none of the researchers running the experiment knew any of the participants before recruiting them for the experiment.
    \item \egcvalue{evaluators include one or more of the authors}: one or more researchers running the experiment was among the participants.
    \item[\checkmark]  \egcvalue{evaluators do not include any of the authors}: none of the researchers running the experiment were among the participants.
    \item \egcvalue{Other} (fewer than 4 of the above apply): we believe you should be able to tick 4 options of the above. If that's not the case, use this box to explain.
\end{enumerate}

\subsubsection*{\qsecbox{Question 3.2.3:  How are evaluators recruited?}}

Evaluators were recruited through a job offer on the University job board and interviewed prior to conducting the experiment.

\subsubsection*{\qsecbox{Question 3.2.4: What training and/or practice are evaluators given before starting on the evaluation itself?}}

First, the evaluators have been introduced to the task of minimal pairs. 
They were then introduced to the dataset under study. 
They learned from an annotation guideline and practices on 15 examples before conducting the whole experiment. 

\subsection*{\qsecbox{Question 3.2.5:  What other characteristics do the evaluators have, known either because these were qualifying criteria, or from information gathered as part of the evaluation?}}

Quebec-French is their native language.

\subsubsection{Experimental design (Questions 3.3.1--3.3.8)}
\subsubsection*{\qsecbox{Question 3.3.1:  Has the experimental design been preregistered? If yes, on which registry?}}
No.

\subsubsection*{\qsecbox{Question 3.3.2: How are responses collected? E.g.\ paper forms, online survey tool, etc.}}

The answers were collected using a customized version of Prodigy\footnote{\url{https://prodi.gy/}}, hosted on Amazon Web Services.

\subsubsection*{\qsecbox{Question 3.3.3:  What quality assurance methods are used? Select all that apply. If none match, select `Other' and describe. In all cases, provide details in the text box under `Other'.}}

\noindent\textit{Check-box options (select all that apply)}:  
\vspace{-.1cm}

\begin{enumerate}[itemsep=0cm,leftmargin=0.5cm,label={\small $\square$}]
    \item[\checkmark] \egcvalue{evaluators are required to be native speakers of the language they evaluate}: mechanisms are in place to ensure all participants are native speakers of the language they evaluate.
    \item \egcvalue{automatic quality checking methods are used during/post evaluation}: evaluations are checked for quality by automatic scripts during or after evaluations, e.g.\ evaluators are given known bad/good outputs to check they're given bad/good scores on MTurk.
    \item[\checkmark] \egcvalue{manual quality checking methods are used during/post evaluation}: evaluations are checked for quality by a manual process  during or after evaluations, e.g.\ scores assigned by evaluators are monitored by researchers conducting the experiment.
    \item \egcvalue{evaluators are excluded if they fail quality checks (often or badly enough)}: there are conditions under which evaluations produced by participants are not included in the final results due to quality issues.
    \item \egcvalue{some evaluations are excluded because of failed quality checks}: there are conditions under which some (but not all) of the evaluations produced by some participants are not included in the final results due to quality issues.
    \item \egcvalue{none of the above}: tick this box if none of the above apply.
    \item \egcvalue{Other (please specify)}: use this box to describe any other quality assurance methods used during or after evaluations, and to provide additional details for any of the options selected above.
\end{enumerate}

\subsubsection*{\qsecbox{Question 3.3.4:  What do evaluators see when carrying out evaluations? Link to screenshot(s) and/or describe the evaluation interface(s).}}

When evaluating, evaluators see the minimal pair.
To reduce any bias toward the annotation pattern, the grammatical and ungrammatical sentence position is sampled.

\subsubsection*{\qsecbox{3.3.5: How free are evaluators regarding when and how quickly to carry out evaluations? Select all that apply. In all cases, provide details in the text box under `Other'.}}

\noindent\textit{Check-box options (select all that apply)}:  
\vspace{-.1cm}

\begin{enumerate}[itemsep=0cm,leftmargin=0.5cm,label={\small $\square$}]
    \item \egcvalue{evaluators have to complete each individual assessment within a set time}: evaluators are timed while carrying out each assessment and cannot complete the assessment once time has run out.
    \item \egcvalue{evaluators have to complete the whole evaluation in one sitting}: partial progress cannot be saved and the evaluation returned to on a later occasion.
    \item[\checkmark] \egcvalue{neither of the above}: Choose this option if neither of the above are the case in the experiment.
    \item \egcvalue{Other (please specify)}: Use this space to describe any other way in which time taken or number of sessions used by evaluators is controlled in the experiment, and to provide additional details for any of the options selected above.
\end{enumerate}

\subsubsection*{\qsecbox{3.3.6: Are evaluators told they can ask questions about the evaluation and/or provide feedback? Select all that apply. In all cases, provide details in the text box under `Other'.}}

\noindent\textit{Check-box options (select all that apply)}:  

\begin{enumerate}[itemsep=0cm,leftmargin=0.5cm,label={\small $\square$}]
    \item[\checkmark] \egcvalue{evaluators are told they can ask any questions during/after receiving initial training/instructions, and before the start of the evaluation}: evaluators are told explicitly that they can ask  questions about the evaluation experiment \textit{before} starting on their assessments, either during or after training.
    \item \egcvalue{evaluators are told they can ask any questions during the evaluation}: evaluators are told explicitly that they can ask  questions about the evaluation experiment \textit{during} their assessments.
    \item \egcvalue{evaluators are asked for feedback and/or comments after the evaluation, e.g.\ via an exit questionnaire or a comment box}: evaluators are explicitly asked to provide feedback and/or comments about the experiment \textit{after} their assessments, either verbally or in written form.
    \item \egcvalue{None of the above}: Choose this option if none of the above are the case in the experiment.
    \item \egcvalue{Other (please specify)}: use this space to describe any other ways you provide for evaluators to ask questions or provide feedback.
\end{enumerate}

\subsubsection*{\qsecbox{3.3.7: What are the experimental conditions in which evaluators carry out the evaluations? If none match, select `Other’ and describe.}}

\noindent\textit{Multiple-choice options (select one)}:  
\begin{enumerate}[itemsep=0cm,leftmargin=0.5cm,label={\LARGE $\circ$}]
    \item[\checkmark] \egcvalue{evaluation carried out by evaluators at a place of their own choosing, e.g.\ online, using a paper form, etc.}: evaluators are given access to the tool or form specified in Question 3.3.2, and subsequently choose where to carry out their evaluations.
    \item \egcvalue{evaluation carried out in a lab, and conditions are the same for each evaluator}: evaluations are carried out in a lab, and conditions in which evaluations are carried out \textit{are} controlled to be the same, i.e.\ the different evaluators all carry out the evaluations in identical conditions of quietness, same type of computer, same room, etc. Note we're not after very fine-grained differences here, such as time of day  or temperature, but the line is difficult to draw, so some judgment is involved here.
    \item \egcvalue{evaluation carried out in a lab, and conditions vary for different evaluators}: choose this option if evaluations are carried out in a lab, but the preceding option does not apply, i.e.\ conditions in which evaluations are carried out are \textit{not} controlled to be the same.
    \item \egcvalue{evaluation carried out in a real-life situation, and conditions are the same for each evaluator}: evaluations are carried out in a real-life situation, i.e.\ one that would occur whether or not the evaluation was carried out (e.g.\ evaluating a dialogue system deployed in a live chat function on a website), and conditions in which evaluations are carried out \textit{are} controlled to be the same. 
    \item \egcvalue{evaluation carried out in a real-life situation, and conditions vary for different evaluators}: choose this option if evaluations are carried out in a real-life situation, but the preceding option does not apply, i.e.\ conditions in which evaluations are carried out are \textit{not} controlled to be the same.
    \item \egcvalue{evaluation carried out outside of the lab, in a situation designed to resemble a real-life situation, and conditions are the same for each evaluator}: evaluations are carried out outside of the lab, in a situation intentionally similar to a real-life situation (but not actually a real-life situation), e.g.\ user-testing a navigation system where the destination is part of the evaluation design, rather than chosen by the user. Conditions in which evaluations are carried out \textit{are} controlled to be the same. 
    \item \egcvalue{evaluation carried out outside of the lab, in a situation designed to resemble a real-life situation, and conditions vary for different evaluators}: choose this option if evaluations are carried out outside of the lab, in a situation intentionally similar to a real-life situation, but the preceding option does not apply, i.e.\ conditions in which evaluations are carried out are \textit{not} controlled to be the same. 
    \item \egcvalue{Other (please specify)}: Use this space to provide additional, or alternative, information about the conditions in which evaluators carry out assessments, not covered by the options above.
\end{enumerate}

\subsubsection*{\qsecbox{3.3.8:  Unless the evaluation is carried out at a place of the evaluators'  own choosing, briefly describe the (range of different) conditions in which evaluators carry out the evaluations.}}
N/A.

\subsection{Quality Criterion \textit{n} -- Definition and Operationalisation}
\label{sec:criteria}

\subsubsection{Quality criterion properties (Questions 4.1.1--4.1.3)}
\subsubsection*{\qsecbox{Question 4.1.1:  What type of quality is assessed by the quality criterion?}}

\noindent\textit{Multiple-choice options (select one)}:  
\begin{enumerate}[itemsep=0cm,leftmargin=0.5cm,label={\LARGE $\circ$}]
    \item[\checkmark] \egcvalue{Correctness}: select this option if it is possible to state,  generally for all outputs,  the conditions under which outputs are maximally correct (hence of maximal quality).  E.g.\ for Grammaticality, outputs are (maximally) correct if they contain no grammatical errors; for Semantic Completeness, outputs are correct if they express all the content in the input.
    \item \egcvalue{Goodness}: select this option if, in contrast to correctness criteria, there is no single, general mechanism for deciding when outputs are maximally good, only for deciding for two outputs which is better and which is worse. E.g.\ for Fluency, even if outputs contain no disfluencies, there may be other ways in which any given output could be more fluent.
    \item \egcvalue{Features}: choose this option if, in terms of property $X$ captured by the criterion, outputs are not generally better if they are more $X$, but instead, depending on evaluation context, more $X$ may be better or less $X$ may be better. E.g.\ outputs can be more specific or less specific, but it’s not the case that outputs are, in the general case, better when they are more specific.
\end{enumerate}

\subsubsection*{\qsecbox{Question 4.1.2:  Which aspect of system outputs is assessed by the quality criterion?}}

\noindent\textit{Multiple-choice options (select one)}:  
\begin{enumerate}[itemsep=0cm,leftmargin=0.5cm,label={\LARGE $\circ$}]
    \item[\checkmark] \egcvalue{Form of output}: choose this option if the criterion assesses the form of outputs alone, e.g.\ Grammaticality is only about the form, a sentence can be grammatical yet be wrong or nonsensical in terms of content.
    \item \egcvalue{Content of output}: choose this option if the criterion assesses the content/meaning of the output alone, e.g.\ Meaning Preservation only assesses output content; two sentences can be considered to have the same meaning, but differ in form.
    \item \egcvalue{Both form and content of output}: choose this option if the criterion assesses outputs as a whole, not just form or just content. E.g.\ Coherence is a property of outputs as a whole, either form or meaning can detract from it.
\end{enumerate}

\subsubsection*{\qsecbox{Question 4.1.3: Is each output assessed for quality in its own right, or with reference to a system-internal or external frame of reference?}}

\noindent\textit{Multiple-choice options (select one)}:  

\begin{enumerate}[itemsep=0cm,leftmargin=0.5cm,label={\LARGE $\circ$}]
    \item[\checkmark] \egcvalue{Quality of output in its own right}: choose this option if output quality is assessed without referring to anything other than the output itself, i.e.\ no  system-internal or external frame of reference. E.g.\ Poeticness is assessed by considering (just) the output and how poetic it is. 
    \item \egcvalue{Quality of output relative to the input}:  choose this option if output quality is assessed  relative to the input. E.g.\ Answerability is the degree to which the output question can be answered from information in the input.
    \item \egcvalue{Quality of output relative to a system-external frame of reference}: choose this option if output quality is assessed with reference to system-external information, such as a knowledge base, a person’s individual writing style, or the performance of an embedding system. E.g.\ Factual Accuracy assesses outputs relative to a source of real-world knowledge.
\end{enumerate}

\subsubsection{Evaluation mode properties (Questions 4.2.1--4.2.3)}

Questions 4.2.1--4.2.3 record properties that are orthogonal to quality criteria, i.e.\ any given quality criterion can in principle be combined with any of the modes (although some combinations are more common than others). 

\subsubsection*{\qsecbox{Question 4.2.1: Does an individual assessment involve an objective or a subjective judgment?}}

\noindent\textit{Multiple-choice options (select one)}:  

\begin{enumerate}[itemsep=0cm,leftmargin=0.5cm,label={\LARGE $\circ$}]
    \item[\checkmark] \egcvalue{Objective}: Examples of objective assessment include any automatically counted or otherwise quantified measurements such as mouse-clicks, occurrences in text, etc. Repeated assessments of the same output with an objective-mode evaluation method always yield the same score/result.
    \item \egcvalue{Subjective}: Subjective assessments involve ratings, opinions and preferences by evaluators. Some criteria lend themselves more readily to subjective assessments, e.g.\ Friendliness of a conversational agent, but an objective measure e.g.\ based on lexical markers is also conceivable.
\end{enumerate}
    
\subsubsection*{\qsecbox{Question 4.2.2: Are outputs assessed in absolute or relative terms?}}

\noindent\textit{Multiple-choice options (select one)}:  
\begin{enumerate}[itemsep=0cm,leftmargin=0.5cm,label={\LARGE $\circ$}]
    \item[\checkmark] \egcvalue{Absolute}: choose this option if evaluators are shown outputs from a single system during each individual assessment. 
    \item \egcvalue{Relative}: choose this option if evaluators are shown outputs from multiple systems at the same time during assessments, typically ranking or preference-judging them.
\end{enumerate}

\subsubsection*{\qsecbox{Question 4.2.3: Is the evaluation intrinsic or extrinsic?}}

\noindent\textit{Multiple-choice options (select one)}:  
\begin{enumerate}[itemsep=0cm,leftmargin=0.5cm,label={\LARGE $\circ$}]
    \item[\checkmark] \egcvalue{Intrinsic}: Choose this option if quality of outputs is assessed \textit{without} considering their \textit{effect} on something external to the system, e.g.\ the performance of an embedding system or of a user at a task.
    \item \egcvalue{Extrinsic}: Choose this option if quality of outputs is assessed in terms of their \textit{effect} on something external to the system such as the performance of an embedding system or of a user at a task.
\end{enumerate}

\subsubsection{Response elicitation (Questions 4.3.1--4.3.11)}

\subsubsection*{\qsecbox{Question 4.3.1: What do you call the quality criterion in explanations/interfaces to evaluators?  Enter `N/A' if criterion not named.}}
N/A

\subsubsection*{\qsecbox{Question 4.3.2:  What definition do you give for the quality criterion in explanations/interfaces to evaluators? Enter `N/A' if no definition given.}}
 
\subsubsection*{\qsecbox{Question 4.3.3:  Size of scale or other rating instrument (i.e.\ how many different possible values there are). Answer should be an integer or `continuous' (if it's not possible to state how many possible responses there are). Enter `N/A' if there is no rating instrument.}}
Binary.

\subsubsection*{\qsecbox{Question 4.3.4: List or range of possible values of the scale or other rating instrument. Enter `N/A', if there is no rating instrument.}} 
$[0, 1]$

\subsubsection*{\qsecbox{Question 4.3.5:  How is the scale or other rating instrument presented to evaluators? If none match, select `Other’ and describe.}}

\noindent\textit{Multiple-choice options (select one)}:  

\begin{enumerate}[itemsep=0cm,leftmargin=0.5cm,label={\LARGE $\circ$}]
    \item \egcvalue{Multiple-choice options}: choose this option if evaluators select exactly one of multiple options.
    \item[\checkmark] \egcvalue{Check-boxes}: choose this option if evaluators select any number of options from multiple given options.
    \item \egcvalue{Slider}: choose this option if evaluators move a pointer on a slider scale to the position corresponding to their assessment.
    \item \egcvalue{N/A (there is no rating instrument)}: choose this option if there is no rating instrument.
    \item \egcvalue{Other (please specify)}: choose this option if there is a rating instrument, but none of the above adequately describe the way you present it to evaluators. Use the text box to describe the rating instrument and link to a screenshot.
\end{enumerate}

\subsubsection*{\qsecbox{Question 4.3.6:  If there is no rating instrument, describe briefly what task the evaluators perform (e.g.\ ranking multiple outputs, finding information, playing a game, etc.), and what information is recorded. Enter `N/A' if there is a rating instrument.}}
N/A.

\subsubsection*{\qsecbox{Question 4.3.7:  What is the verbatim question, prompt or instruction given to evaluators (visible to them during each individual assessment)?}}

Here is the verbatim question and instruction in French to evaluators, we also present an automatic translation.

\textbf{\textit{Sélectionnez la phrase que vous jugez bien écrite parmi les deux phrases disponibles.} }

Here is the automatic English translation of the verbatim question and instructions for evaluators.

\textbf{Select the sentence you think is well written from the two available.}

\subsubsection*{\qsecbox{Question 4.3.8:  Form of response elicitation. If none match, select `Other' and describe.}}
\noindent\textit{Multiple-choice options (select one)}:\footnote{Explanations adapted from \citet{howcroft2019typology}.}

\begin{enumerate}[itemsep=0cm,leftmargin=0.5cm,label={\LARGE $\circ$}]
    \item \egcvalue{(dis)agreement with quality statement}: Participants specify the degree to which they agree with a given quality statement by indicating their agreement on a rating instrument. The rating instrument is labelled with degrees of agreement and can additionally have numerical labels.  E.g.\ \textit{This text is fluent --- 1=strongly disagree...5=strongly agree}.
    \item \egcvalue{direct quality estimation}: Participants are asked to provide a rating using a rating instrument, which typically (but not always) mentions the quality criterion explicitly. E.g.\ \textit{How fluent is this text? --- 1=not at all fluent...5=very fluent}.
    \item[\checkmark] \egcvalue{relative quality estimation (including ranking)}: Participants evaluate two or more items in terms of which is better.
    E.g.\ \textit{Rank these texts in terms of fluency}; \textit{Which of these texts is more fluent?}; \textit{Which of these items do you prefer?}.
    \item \egcvalue{counting occurrences in text}: Evaluators are asked to count how many times some type of phenomenon occurs, e.g.\ the number of facts contained in the output that are inconsistent with the input.
    \item \egcvalue{qualitative feedback (e.g.\ via comments entered in a text box)}: Typically, these are responses to open-ended questions in a survey or interview.
    \item \egcvalue{evaluation through post-editing/annotation}: Choose this option if the evaluators' task consists of editing or inserting annotations in text. E.g.\ evaluators may perform error correction and edits are then automatically measured to yield a numerical score.
    \item \egcvalue{output classification or labelling}: Choose this option if evaluators assign outputs to categories. E.g.\ \textit{What is the overall sentiment of this piece of text? --- Positive/neutral/negative.}
    \item \egcvalue{user-text interaction measurements}: choose this option if participants in the evaluation experiment interact with a text in some way, and measurements are taken of their interaction. E.g.\ reading speed, eye movement tracking, comprehension questions, etc. Excludes situations where participants are given a task to solve and their performance is measured which comes under the next option.
    \item \egcvalue{task performance measurements}: choose this option if participants in the evaluation experiment are given a task to perform, and measurements are taken of their performance at the task.  E.g.\ task is finding information, and task performance measurement is task completion speed and success rate.
    \item \egcvalue{user-system interaction measurements}: choose this option if participants in the evaluation experiment interact with a system in some way, while measurements are taken of their interaction. E.g.\ duration of interaction, hyperlinks followed, number of likes, or completed sales.
    \item \egcvalue{Other (please specify)}: Use the text box to describe the form of response elicitation used in assessing the quality criterion if it doesn't fall in any of the above categories.
\end{enumerate}

\subsubsection*{\qsecbox{Question 4.3.9:  How are raw responses from participants aggregated or otherwise processed to obtain reported scores for this quality criterion? State if no scores reported.}}

Macro averages are computed from numerical scores to provide a summary.

\subsubsection*{\qsecbox{Question 4.3.10:  Method(s) used for determining effect size and significance of findings for this quality criterion.}}

\noindent\textit{What to enter in the text box}: A list of  methods used for calculating the effect size and significance of any results, both as reported in the paper given in Question 1.1, for this quality criterion. If none calculated, state `None'.

\bigskip
\noindent None.

\vspace{-.3cm}
\subsubsection*{\qsecbox{Question 4.3.11:  Has the inter-annotator and intra-annotator agreement between evaluators for this quality criterion been measured? If yes, what method was used, and what are the agreement scores?}}

Worker Agreement With Aggregate (WAWA) coefficient \cite{ning-etal-2018-joint} is used to measure inter-annotator agreement. WAWA coefficients are detailled in \autoref{tab:annotators_res}.

\subsection{Ethics}\label{sec:ethics}

\subsubsection*{\qsecbox{Question 5.1: Has the evaluation experiment this sheet is being completed for, or the larger study it is part of, been approved by a research ethics committee? If yes, which research ethics committee?}}
No.
\vspace{-0.4cm}

\subsubsection*{\qsecbox{Question 5.2: Do any of the system outputs (or human-authored stand-ins) evaluated, or do any of the responses collected, in the experiment contain personal data (as defined in GDPR Art. 4, §1: https://gdpr.eu/article-4-definitions/)? If yes, describe data and state how addressed.}}
No.
\vspace{-0.4cm}

\subsubsection*{\qsecbox{Question 5.3: Do any of the system outputs (or human-authored stand-ins) evaluated, or do any of the responses collected, in the experiment contain special category information (as defined in GDPR Art. 9, §1: https://gdpr.eu/article-9-processing-special-categories-of-personal-data-prohibited/)? If yes, describe data and state how addressed.}}
No.
\vspace{-0.4cm}

\subsubsection*{\qsecbox{Question 5.4:  Have any impact assessments been carried out for the evaluation experiment, and/or any data collected/evaluated in connection with it? If yes, summarise approach(es) and outcomes.}}
No.

\clearpage

\section{Selected LLM Details}
\label{an:selectedllmdetails}
We present in \autoref{tab:selectedllm} the comprehensive suite of open-source LLMs benchmarked in our study, detailing their origins and respective sizes. 
The selection was curated to cover a wide spectrum of parameter counts, and to include those with specializations in French ($\Upsilon$) or reasoning ($\Gamma$).
All LLMs are downloaded from the \href{https://huggingface.co/models}{HuggingFace Model repository} \citep{wolf2020huggingfaces} using default parameters.

\begin{table*}
    \resizebox{\textwidth}{!}{%
    \centering
    \begin{tabular}{llcllc}
        \toprule
        LLM & Source & Size & LLM & Source & Size\\
        \midrule
        \texttt{Aya-23-8b} & \citet{aryabumi2024aya} & 8B &\texttt{Lucie-7b-it-human-data} ($\Upsilon$) & \citet{openllm2025lucie} & 6.71B \\
        \texttt{Aya-expanse-8b} & \citet{dang2024ayaexpansecombiningresearch} & 8B &        \texttt{Lucie-7b-it} ($\Upsilon$) & \citet{openllm2025lucie} & 6.71B \\
        \texttt{BERT-base-French-europeana} ($\Upsilon$) & \citet{europeana} & 110.6M &        \texttt{Lucie-7b} ($\Upsilon$) & \citet{openllm2025lucie} & 6.71B \\
        \texttt{Bloom-1b1} & \citet{workshop2022bloom} & 1B &        \texttt{Meta-Llama-$3.1$-70b-it} ($\Gamma$) & \citet{grattafiori2024llama} & 70.6B \\
        \texttt{Bloom-1b7} & \citet{workshop2022bloom} & 1.7B &        \texttt{Meta-Llama-$3.1$-70b} ($\Gamma$) & \citet{grattafiori2024llama} & 70.6B \\
        \texttt{Bloom-560m} & \citet{workshop2022bloom} & 559.2M &        \texttt{Meta-Llama-$3.1$-8b-it} ($\Gamma$) & \citet{grattafiori2024llama} & 8B \\
        \texttt{Bloom-7b1} & \citet{workshop2022bloom} & 7B &        \texttt{Meta-Llama-$3.1$-8b} ($\Gamma$) & \citet{grattafiori2024llama} & 8B \\
        \texttt{Bloomz-1b1} & \citet{muennighoff2023crosslingual} & 1B &        \texttt{Ministral-8b-it} & \citet{rastogi2025magistral} & 8B \\
        \texttt{Bloomz-560m} & \citet{muennighoff2023crosslingual} & 559.2M &        \texttt{Mistral-7b-it} & \citet{rastogi2025magistral} & 7.2B \\
        \texttt{CamemBERT-base} ($\Upsilon$) & \citet{martin2020camembert} & 110.6M &        \texttt{Mistral-7b} & \citet{rastogi2025magistral} & 7.2B \\
        \texttt{CamemBERT-large} ($\Upsilon$) & \citet{martin2020camembert} & 336.6M &        \texttt{Mistral-nemo-it} & \citet{rastogi2025magistral} & 12.2B \\
        \texttt{Chocolatine-14b-it} ($\Upsilon$) & \citet{chocolatine} & 14B &        \texttt{Mistral-small-it} & \citet{rastogi2025magistral} & 22.2B \\
        \texttt{Chocolatine-2-14b-it} ($\Upsilon$) & \citet{chocolatinev2} & 14.8B &        \texttt{Mixtral-8x7b-it} & \citet{rastogi2025magistral} & 46.7B \\
        \texttt{Claire-7b-FR-it} ($\Upsilon$) & \citet{openllmfrance} & 6.92B &        \texttt{Mixtral-8x7b} & \citet{rastogi2025magistral} & 46.7B \\
        \texttt{DeepSeek-R1-distill-Llama-8b} ($\Gamma$) & \citet{deepseekai2025deepseekr1incentivizingreasoningcapability} & 8.03B &        \texttt{Phi-$3.5$-mini-it} & \citet{abdin2024phi3technicalreporthighly} & 3.8B \\
        \texttt{DeepSeek-R1-distill-Qwen-14b} ($\Gamma$) & \citet{deepseekai2025deepseekr1incentivizingreasoningcapability} & 14.8B &      \texttt{Phi-4} & \citet{abdin2024phi} & 14.7B \\
        \texttt{DeepSeek-R1-distill-Qwen-32b} ($\Gamma$) & \citet{deepseekai2025deepseekr1incentivizingreasoningcapability} & 32.8B &        \texttt{QwQ-32b} ($\Gamma$) & \citet{qwq32b} ($\Gamma$) & 32.8B \\
        \texttt{DeepSeek-R1-distill-Qwen-7b} ($\Gamma$) &  \citet{deepseekai2025deepseekr1incentivizingreasoningcapability}& 7.62B &        \texttt{Qwen$2.5$-$0.5$b-it} ($\Gamma$) & \citet{hui2024qwen2}  & 494M \\
        \texttt{Deepthink-reasoning-14b} ($\Gamma$) & \citet{deepthink2} & 14.8B &        \texttt{Qwen$2.5$-$0.5$b} & \citet{hui2024qwen2} & 494M \\
        \texttt{Deepthink-reasoning-7b} ($\Gamma$) & \citet{deepthink1} & 7.62B &        \texttt{Qwen$2.5$-$1.5$b-it} & \citet{hui2024qwen2}  & 1.5B \\
        \texttt{FLAN-T5-base} & \citet{https://doi.org/10.48550/arxiv.2210.11416} & 247.5M &        \texttt{Qwen$2.5$-$1.5$b} & \citet{hui2024qwen2}  & 1.5B \\
        \texttt{FLAN-T5-large} & \citet{https://doi.org/10.48550/arxiv.2210.11416} & 783.1M &        \texttt{Qwen$2.5$-14b-it} & \citet{hui2024qwen2}  & 14.7B \\
        \texttt{FLAN-T5-small} & \citet{https://doi.org/10.48550/arxiv.2210.11416} & 76.9M &        \texttt{Qwen$2.5$-14b} & \citet{hui2024qwen2}  & 14.7B \\
        \texttt{FLAN-T5-xl} & \citet{https://doi.org/10.48550/arxiv.2210.11416} & 2.8B &        \texttt{Qwen$2.5$-32b-it} & \citet{hui2024qwen2}  & 32.8B \\
        \texttt{FLAN-T5-xxl} & \citet{https://doi.org/10.48550/arxiv.2210.11416} & 11.1B &        \texttt{Qwen$2.5$-32b} & \citet{hui2024qwen2}  & 32.8B \\
        \texttt{French-Alpaca-Llama3-8b-it} ($\Upsilon$, $\Gamma$) & \citet{alpaca} & 8.03B &  \texttt{Qwen$2.5$-3b-it} & \citet{hui2024qwen2}  & 3B \\
        \texttt{Gemma-2-27b-it} ($\Gamma$) & \citet{mesnard2024gemma} & 27.2B   &        \texttt{Qwen$2.5$-3b} & \citet{hui2024qwen2}  & 3B \\
        \texttt{Gemma-2-27b} ($\Gamma$) & \citet{mesnard2024gemma} & 27.2B   &         \texttt{Qwen$2.5$-72b-it} & \citet{hui2024qwen2}  & 72.7B \\
         \texttt{Gemma-2-2b-it} ($\Gamma$) & \citet{mesnard2024gemma} & 27.2B &        \texttt{Qwen$2.5$-72b} & \citet{hui2024qwen2}  & 72.7B \\
        \texttt{Gemma-2-2b} ($\Gamma$)& \citet{mesnard2024gemma} & 2.6B &        \texttt{Qwen$2.5$-7b-it} & \citet{hui2024qwen2}  & 7.6B \\
         \texttt{Gemma-2-9b-it} ($\Gamma$) & \citet{mesnard2024gemma} & 9B &        \texttt{Qwen$2.5$-7b} & \citet{hui2024qwen2}  & 7.6B \\
         \texttt{Gemma-2-9b} ($\Gamma$)& \citet{mesnard2024gemma} & 9.2B &        \texttt{Reka-flash-3} ($\Gamma$) & \citet{reka} & 20.9B \\
         \texttt{Llama-$3.2$-1b-it} ($\Gamma$) & \citet{grattafiori2024llama} & 1.2B &        \texttt{S1.1-32b} ($\Gamma$) & \citet{s11} & 32.8B \\
        \texttt{Llama-$3.2$-1b} ($\Gamma$)  & \citet{grattafiori2024llama} & 1.2B &         \texttt{SmolLM2-$1.7$b-it} & \citet{allal2025smollm2smolgoesbig}  & 1.7B \\
       \texttt{Llama-$3.2$-3b-it} ($\Gamma$) & \citet{grattafiori2024llama} & 3.21B &          \texttt{SmolLM2-$1.7$b} & \citet{allal2025smollm2smolgoesbig}  & 1.7B \\
       \texttt{Llama-$3.2$-3b} ($\Gamma$) & \citet{grattafiori2024llama} & 3.21B &           \texttt{SmolLM2-135m-it} & \citet{allal2025smollm2smolgoesbig}  & 134.5M \\  
        &&& \texttt{SmolLM2-135m} &\citet{allal2025smollm2smolgoesbig}  & 134.5M \\
        &&& \texttt{SmolLM2-360m-it} & \citet{allal2025smollm2smolgoesbig}  & 361.8M \\
        &&& \texttt{SmolLM2-360m} & \citet{allal2025smollm2smolgoesbig}  & 361.8M \\
        &&&\texttt{XLM-roBERTa-base} & \citet{DBLP:journals/corr/abs-1911-02116} & 278.2M \\
         &&& \texttt{XLM-roBERTa-large} & \citet{DBLP:journals/corr/abs-1911-02116} & 560.1M \\
        \bottomrule
    \end{tabular}%
    }
    \caption{The selected open-source LLM used in our work, along with their source and size. \guillemet{$\Upsilon$} are model that have a specialization in French, while $\Gamma$ are model marketed as reasoning LLM.}
    \label{tab:selectedllm}
\end{table*}

\section{Complete Results}
\label{ann:completeresults}

In this section, we present the complete results of our analysis.
\autoref{tab:resultscomp} present the complete results of our \llm{} evaluated LLMs, while \autoref{tab:resultsdiffcomp} present the difference of each LLM performance agains our \texttt{Human} baseline.

\begin{table*}
    \centering
    \resizebox{\textwidth}{!}{%
    \begin{tabular}{lRRRRRRRRRRRRRRRRRRRR}
        \toprule
        LLM & \multicolumn{1}{c}{\texttt{1}} & 
        \multicolumn{1}{c}{\texttt{2}} & 
        \multicolumn{1}{c}{\texttt{3}} & 
        \multicolumn{1}{c}{\texttt{4}} & 
        \multicolumn{1}{c}{\texttt{5}} & 
        \multicolumn{1}{c}{\texttt{6}} & 
        \multicolumn{1}{c}{\texttt{7}} & 
        \multicolumn{1}{c}{\texttt{8}} &
        \multicolumn{1}{c}{\texttt{9}} & 
        \multicolumn{1}{c}{\texttt{10}} & 
        \multicolumn{1}{c}{\texttt{11}} &
        \multicolumn{1}{c}{\texttt{12}} & 
        \multicolumn{1}{c}{\texttt{13}} &
        \multicolumn{1}{c}{\texttt{14}} &
        \multicolumn{1}{c}{\texttt{15}} & 
        \multicolumn{1}{c}{\texttt{16}} & 
        \multicolumn{1}{c}{\texttt{17}} &
        \multicolumn{1}{c}{\texttt{18}} &
        \multicolumn{1}{c}{\texttt{19}} &
        \multicolumn{1}{c}{\texttt{20}} \\
        \midrule
        \texttt{Aya-23-8b} & 84.54 & 81.05 & 84.54 & 92.71 & 96.97 & 69.07 & 80.70 & 70.75 & 61.95 & 82.42 & 80.00 & 90.00 & 72.92 & 88.66 & 76.67 & 73.33 & 83.33 & 65.00 & 70.00 & 20.63 \\
        \texttt{Aya-expanse-8b} & 82.47 & 83.16 & 83.51 & 89.58 & 96.97 & 72.16 & 82.46 & 75.47 & 60.18 & 87.91 & 84.00 & 94.00 & 77.08 & 91.75 & 73.33 & 66.67 & 76.67 & 71.67 & 70.00 & 25.40 \\
        \texttt{BERT-base-French-europeana} ($\Upsilon$) & 63.92 & 68.42 & 47.42 & 77.08 & 54.55 & 49.48 & 83.33 & 61.32 & 56.64 & 48.35 & 68.00 & 74.00 & 40.62 & 27.84 & 90.00 & 88.33 & 88.33 & 91.67 & 71.67 & 1.59 \\
        \texttt{Bloom-1b1} & 88.66 & 95.79 & 97.94 & 91.67 & 97.98 & 93.81 & 87.72 & 82.08 & 65.49 & 91.21 & 91.00 & 98.00 & 81.25 & 97.94 & 81.67 & 90.00 & 80.00 & 81.67 & 85.00 & 25.40 \\
        \texttt{Bloom-1b7} & 88.66 & 97.89 & 97.94 & 92.71 & 96.97 & 94.85 & 91.23 & 85.85 & 69.03 & 93.41 & 91.00 & 100.00 & 80.21 & 97.94 & 86.67 & 96.67 & 80.00 & 81.67 & 83.33 & 28.57 \\
        \texttt{Bloom-560m} & 89.69 & 90.53 & 98.97 & 92.71 & 95.96 & 91.75 & 88.60 & 85.85 & 61.95 & 91.21 & 90.00 & 97.00 & 84.38 & 97.94 & 81.67 & 90.00 & 76.67 & 76.67 & 80.00 & 20.63 \\
        \texttt{Bloom-7b1} & 91.75 & 94.74 & 96.91 & 91.67 & 95.96 & 92.78 & 93.86 & 84.91 & 69.03 & 92.31 & 95.00 & 100.00 & 85.42 & 97.94 & 80.00 & 96.67 & 83.33 & 81.67 & 83.33 & 42.86 \\
        \texttt{Bloomz-1b1} & 83.51 & 95.79 & 97.94 & 91.67 & 96.97 & 94.85 & 90.35 & 85.85 & 61.95 & 89.01 & 91.00 & 99.00 & 82.29 & 96.91 & 78.33 & 86.67 & 76.67 & 81.67 & 85.00 & 39.68 \\
        \texttt{Bloomz-560m} & 85.57 & 90.53 & 93.81 & 91.67 & 96.97 & 91.75 & 86.84 & 83.96 & 60.18 & 91.21 & 86.00 & 97.00 & 81.25 & 97.94 & 78.33 & 88.33 & 73.33 & 71.67 & 85.00 & 23.81 \\
        \texttt{CamemBERT-base} ($\Upsilon$) & 59.79 & 44.21 & 45.36 & 67.71 & 56.57 & 42.27 & 81.58 & 49.06 & 35.40 & 51.65 & 60.00 & 64.00 & 43.75 & 28.87 & 91.67 & 100.00 & 90.00 & 96.67 & 75.00 & 1.59 \\
        \texttt{CamemBERT-large} ($\Upsilon$) & 48.45 & 38.95 & 42.27 & 62.50 & 30.30 & 49.48 & 82.46 & 55.66 & 40.71 & 52.75 & 65.00 & 52.00 & 35.42 & 22.68 & 93.33 & 100.00 & 90.00 & 95.00 & 65.00 & 0.00 \\
        \texttt{Chocolatine-14b-it} ($\Upsilon$) & 90.72 & 97.89 & 93.81 & 95.83 & 95.96 & 94.85 & 92.11 & 88.68 & 76.11 & 90.11 & 94.00 & 99.00 & 84.38 & 81.44 & 86.67 & 88.33 & 88.33 & 76.67 & 78.33 & 44.44 \\
        \texttt{Chocolatine-14b-it} ($\Upsilon$) & 90.72 & 97.89 & 93.81 & 95.83 & 95.96 & 94.85 & 92.11 & 88.68 & 76.11 & 90.11 & 94.00 & 99.00 & 84.38 & 81.44 & 86.67 & 88.33 & 88.33 & 76.67 & 78.33 & 44.44 \\
        \texttt{Chocolatine-2-14b-it} ($\Upsilon$) & 91.75 & 98.95 & 95.88 & 97.92 & 96.97 & 96.91 & 95.61 & 88.68 & 71.68 & 92.31 & 90.00 & 98.00 & 84.38 & 89.69 & 83.33 & 95.00 & 81.67 & 81.67 & 73.33 & 42.86 \\
        \texttt{Claire-7b-fr-it} ($\Upsilon$) & 92.78 & 97.89 & 97.94 & 97.92 & 97.98 & 96.91 & 95.61 & 91.51 & 69.03 & 96.70 & 92.00 & 99.00 & 87.50 & 81.44 & 91.67 & 95.00 & 88.33 & 81.67 & 76.67 & 38.10 \\
        \texttt{DeepSeek-R1-distill-Llama-8b} ($\Gamma$) & 91.75 & 91.58 & 95.88 & 93.75 & 94.95 & 86.60 & 76.32 & 88.68 & 61.95 & 86.81 & 90.00 & 96.00 & 81.25 & 81.44 & 78.33 & 93.33 & 80.00 & 75.00 & 65.00 & 33.33 \\
        \texttt{DeepSeek-R1-distill-Qwen-14b} ($\Gamma$) & 91.75 & 95.79 & 97.94 & 96.88 & 97.98 & 95.88 & 86.84 & 92.45 & 69.03 & 85.71 & 93.00 & 98.00 & 77.08 & 88.66 & 83.33 & 93.33 & 83.33 & 86.67 & 73.33 & 47.62 \\
        \texttt{DeepSeek-R1-distill-Qwen-32b} ($\Gamma$) & 92.78 & 95.79 & 96.91 & 92.71 & 94.95 & 93.81 & 92.11 & 88.68 & 68.14 & 90.11 & 92.00 & 98.00 & 84.38 & 85.57 & 85.00 & 98.33 & 81.67 & 88.33 & 66.67 & 57.14 \\
        \texttt{DeepSeek-R1-distill-Qwen-7b} ($\Gamma$) & 80.41 & 80.00 & 95.88 & 86.46 & 94.95 & 85.57 & 76.32 & 73.58 & 55.75 & 71.43 & 83.00 & 88.00 & 69.79 & 81.44 & 66.67 & 85.00 & 78.33 & 61.67 & 50.00 & 36.51 \\
        \texttt{Deepthink-reasoning-14b} ($\Gamma$) & 92.78 & 96.84 & 97.94 & 94.79 & 95.96 & 95.88 & 94.74 & 91.51 & 71.68 & 92.31 & 92.00 & 98.00 & 83.33 & 90.72 & 80.00 & 93.33 & 85.00 & 86.67 & 76.67 & 38.10 \\
        \texttt{Deepthink-reasoning-7b} ($\Gamma$) & 93.81 & 98.95 & 94.85 & 97.92 & 96.97 & 96.91 & 94.74 & 89.62 & 67.26 & 91.21 & 90.00 & 99.00 & 81.25 & 88.66 & 86.67 & 93.33 & 88.33 & 86.67 & 73.33 & 53.97 \\
        \texttt{FLAN-T5-base} & 79.38 & 67.37 & 70.10 & 78.12 & 81.82 & 58.76 & 65.79 & 70.75 & 57.52 & 60.44 & 70.00 & 82.00 & 58.33 & 53.61 & 10.00 & 20.00 & 18.33 & 21.67 & 73.33 & 23.81 \\
        \texttt{FLAN-T5-large} & 78.35 & 64.21 & 62.89 & 54.17 & 89.90 & 73.20 & 78.95 & 63.21 & 46.02 & 53.85 & 78.00 & 75.00 & 53.12 & 70.10 & 18.33 & 30.00 & 20.00 & 23.33 & 66.67 & 52.38 \\
        \texttt{FLAN-T5-small} & 59.79 & 50.53 & 47.42 & 61.46 & 67.68 & 55.67 & 71.93 & 66.04 & 47.79 & 39.56 & 57.00 & 57.00 & 51.04 & 37.11 & 25.00 & 26.67 & 25.00 & 35.00 & 65.00 & 47.62 \\
        \texttt{FLAN-T5-xl} & 83.51 & 78.95 & 68.04 & 58.33 & 92.93 & 53.61 & 81.58 & 58.49 & 44.25 & 58.24 & 82.00 & 72.00 & 56.25 & 74.23 & 25.00 & 43.33 & 45.00 & 60.00 & 63.33 & 33.33 \\
        \texttt{FLAN-T5-xxl} & 74.23 & 68.42 & 52.58 & 60.42 & 79.80 & 67.01 & 73.68 & 65.09 & 51.33 & 48.35 & 77.00 & 74.00 & 53.12 & 57.73 & 55.00 & 68.33 & 45.00 & 65.00 & 63.33 & 31.75 \\
        \texttt{French-Alpaca-Llama3-8b-it} ($\Upsilon$, $\Gamma$) & 92.78 & 90.53 & 95.88 & 93.75 & 96.97 & 92.78 & 91.23 & 89.62 & 74.34 & 87.91 & 90.00 & 96.00 & 85.42 & 85.57 & 80.00 & 96.67 & 88.33 & 86.67 & 75.00 & 28.57 \\
        \texttt{Gemma-2-27b-it} ($\Gamma$) & 88.66 & 87.37 & 96.91 & 95.83 & 96.97 & 91.75 & 91.23 & 85.85 & 61.95 & 94.51 & 88.00 & 93.00 & 77.08 & 84.54 & 91.67 & 98.33 & 93.33 & 91.67 & 83.33 & 33.33 \\
        \texttt{Gemma-2-27b} ($\Gamma$) & 84.54 & 97.89 & 96.91 & 97.92 & 96.97 & 96.91 & 97.37 & 83.96 & 66.37 & 97.80 & 92.00 & 98.00 & 81.25 & 83.51 & 95.00 & 98.33 & 96.67 & 93.33 & 81.67 & 38.10 \\
        \texttt{Gemma-2-2b-it} ($\Gamma$) & 83.51 & 91.58 & 95.88 & 95.83 & 100.00 & 88.66 & 88.60 & 86.79 & 65.49 & 91.21 & 85.00 & 93.00 & 83.33 & 79.38 & 91.67 & 96.67 & 90.00 & 83.33 & 70.00 & 22.22 \\
        \texttt{Gemma-2-2b} ($\Gamma$) & 85.57 & 96.84 & 95.88 & 93.75 & 98.99 & 89.69 & 92.11 & 87.74 & 65.49 & 91.21 & 88.00 & 96.00 & 79.17 & 76.29 & 95.00 & 96.67 & 91.67 & 90.00 & 75.00 & 28.57 \\
        \texttt{Gemma-2-9b-it} ($\Gamma$) & 87.63 & 92.63 & 96.91 & 93.75 & 98.99 & 87.63 & 85.96 & 85.85 & 67.26 & 89.01 & 91.00 & 94.00 & 78.12 & 82.47 & 98.33 & 95.00 & 93.33 & 93.33 & 80.00 & 41.27 \\
        \texttt{Gemma-2-9b} ($\Gamma$) & 88.66 & 95.79 & 94.85 & 95.83 & 97.98 & 92.78 & 94.74 & 89.62 & 67.26 & 93.41 & 90.00 & 98.00 & 81.25 & 81.44 & 98.33 & 100.00 & 95.00 & 91.67 & 81.67 & 30.16 \\
        \texttt{Llama-$3.2$-1b-it} ($\Gamma$) & 87.63 & 88.42 & 92.78 & 94.79 & 98.99 & 92.78 & 82.46 & 85.85 & 69.03 & 84.62 & 88.00 & 95.00 & 82.29 & 82.47 & 76.67 & 88.33 & 86.67 & 60.00 & 66.67 & 19.05 \\
        \texttt{Llama-$3.2$-1b} ($\Gamma$) & 88.66 & 89.47 & 94.85 & 91.67 & 98.99 & 90.72 & 90.35 & 84.91 & 73.45 & 87.91 & 87.00 & 93.00 & 82.29 & 85.57 & 91.67 & 91.67 & 90.00 & 73.33 & 73.33 & 19.05 \\
        \texttt{Llama-$3.2$-3b-it} ($\Gamma$) & 93.81 & 91.58 & 95.88 & 93.75 & 97.98 & 93.81 & 91.23 & 87.74 & 70.80 & 87.91 & 88.00 & 94.00 & 82.29 & 88.66 & 91.67 & 95.00 & 86.67 & 80.00 & 76.67 & 22.22 \\
        \texttt{Llama-$3.2$-3b} ($\Gamma$) & 92.78 & 93.68 & 96.91 & 93.75 & 100.00 & 91.75 & 93.86 & 87.74 & 68.14 & 93.41 & 89.00 & 96.00 & 88.54 & 89.69 & 90.00 & 95.00 & 93.33 & 93.33 & 73.33 & 20.63 \\
        \texttt{Lucie-7b-it-human-data} ($\Upsilon$) & 76.29 & 83.16 & 94.85 & 89.58 & 96.97 & 92.78 & 91.23 & 86.79 & 57.52 & 90.11 & 85.00 & 94.00 & 76.04 & 75.26 & 86.67 & 98.33 & 91.67 & 96.67 & 71.67 & 15.87 \\
        \texttt{Lucie-7b-it} ($\Upsilon$) & 79.38 & 97.89 & 95.88 & 92.71 & 100.00 & 92.78 & 92.11 & 86.79 & 66.37 & 94.51 & 94.00 & 98.00 & 81.25 & 83.51 & 96.67 & 100.00 & 91.67 & 90.00 & 78.33 & 22.22 \\
        \texttt{Lucie-7b} ($\Upsilon$) & 89.69 & 98.95 & 95.88 & 96.88 & 98.99 & 94.85 & 96.49 & 92.45 & 66.37 & 96.70 & 94.00 & 100.00 & 83.33 & 89.69 & 90.00 & 98.33 & 93.33 & 95.00 & 75.00 & 39.68 \\
        \texttt{Meta-Llama-$3.1$-70b-it} ($\Gamma$) & 91.75 & 94.74 & 93.81 & 95.83 & 96.97 & 95.88 & 89.47 & 88.68 & 70.80 & 91.21 & 95.00 & 93.00 & 83.33 & 90.72 & 95.00 & 100.00 & 88.33 & 91.67 & 73.33 & 41.27 \\
        \texttt{Meta-Llama-$3.1$-70b} ($\Gamma$) & 91.75 & 96.84 & 94.85 & 93.75 & 97.98 & 94.85 & 94.74 & 94.34 & 75.22 & 93.41 & 92.00 & 96.00 & 82.29 & 86.60 & 96.67 & 100.00 & 86.67 & 86.67 & 68.33 & 49.21 \\
        \texttt{Meta-Llama-$3.1$-8b-it} ($\Gamma$) & 91.75 & 94.74 & 96.91 & 96.88 & 97.98 & 93.81 & 93.86 & 87.74 & 78.76 & 93.41 & 92.00 & 96.00 & 84.38 & 87.63 & 90.00 & 98.33 & 91.67 & 83.33 & 71.67 & 30.16 \\
        \texttt{Meta-Llama-$3.1$-8b} ($\Gamma$) & 90.72 & 94.74 & 97.94 & 95.83 & 96.97 & 92.78 & 94.74 & 87.74 & 73.45 & 92.31 & 89.00 & 96.00 & 82.29 & 85.57 & 85.00 & 100.00 & 95.00 & 80.00 & 75.00 & 31.75 \\
        \texttt{Ministral-8b-it} & 85.57 & 92.63 & 96.91 & 96.88 & 95.96 & 95.88 & 94.74 & 87.74 & 58.41 & 91.21 & 93.00 & 98.00 & 81.25 & 91.75 & 91.67 & 98.33 & 91.67 & 93.33 & 85.00 & 30.16 \\
        \texttt{Mistral-7b-it} & 90.72 & 94.74 & 96.91 & 93.75 & 95.96 & 89.69 & 87.72 & 88.68 & 68.14 & 91.21 & 90.00 & 96.00 & 84.38 & 79.38 & 88.33 & 95.00 & 90.00 & 91.67 & 76.67 & 26.98 \\
        \texttt{Mistral-7b} & 86.60 & 91.58 & 95.88 & 93.75 & 95.96 & 92.78 & 90.35 & 83.02 & 69.91 & 91.21 & 91.00 & 95.00 & 82.29 & 82.47 & 88.33 & 93.33 & 90.00 & 85.00 & 73.33 & 26.98 \\
        \texttt{Mistral-nemo-it} & 85.57 & 93.68 & 98.97 & 98.96 & 97.98 & 95.88 & 93.86 & 89.62 & 59.29 & 92.31 & 92.00 & 97.00 & 81.25 & 92.78 & 90.00 & 98.33 & 90.00 & 88.33 & 80.00 & 36.51 \\
        \texttt{Mistral-small-it} & 91.75 & 95.79 & 95.88 & 95.83 & 97.98 & 91.75 & 96.49 & 89.62 & 66.37 & 92.31 & 94.00 & 98.00 & 84.38 & 86.60 & 86.67 & 98.33 & 86.67 & 90.00 & 80.00 & 39.68 \\
        \texttt{Mixtral-8x7b-it} & 89.69 & 95.79 & 93.81 & 97.92 & 97.98 & 95.88 & 99.12 & 89.62 & 66.37 & 93.41 & 94.00 & 99.00 & 81.25 & 86.60 & 93.33 & 96.67 & 90.00 & 86.67 & 76.67 & 42.86 \\
        \texttt{Mixtral-8x7b-it} & 90.72 & 96.84 & 93.81 & 97.92 & 94.95 & 95.88 & 99.12 & 91.51 & 66.37 & 92.31 & 94.00 & 99.00 & 81.25 & 87.63 & 91.67 & 96.67 & 85.00 & 86.67 & 78.33 & 44.44 \\
        \texttt{Mixtral-8x7b} & 88.66 & 94.74 & 94.85 & 96.88 & 95.96 & 96.91 & 98.25 & 85.85 & 65.49 & 91.21 & 90.00 & 100.00 & 81.25 & 90.72 & 88.33 & 98.33 & 88.33 & 85.00 & 75.00 & 36.51 \\
        \texttt{Mixtral-8x7b} & 91.75 & 94.74 & 95.88 & 97.92 & 96.97 & 95.88 & 99.12 & 89.62 & 65.49 & 93.41 & 93.00 & 100.00 & 82.29 & 88.66 & 90.00 & 98.33 & 88.33 & 90.00 & 78.33 & 46.03 \\
        \texttt{Phi-3-mini-4k-it} & 88.66 & 93.68 & 96.91 & 96.88 & 95.96 & 91.75 & 89.47 & 88.68 & 69.03 & 85.71 & 92.00 & 97.00 & 82.29 & 82.47 & 85.00 & 93.33 & 83.33 & 78.33 & 73.33 & 41.27 \\
        \texttt{Phi-$3.5$-mini-it} & 88.66 & 93.68 & 97.94 & 96.88 & 98.99 & 93.81 & 93.86 & 89.62 & 74.34 & 85.71 & 90.00 & 97.00 & 83.33 & 80.41 & 90.00 & 88.33 & 90.00 & 63.33 & 70.00 & 36.51 \\
        \texttt{Phi-4} ($\Gamma$) & 92.78 & 98.95 & 96.91 & 96.88 & 96.97 & 97.94 & 97.37 & 88.68 & 71.68 & 95.60 & 93.00 & 99.00 & 82.29 & 92.78 & 83.33 & 96.67 & 86.67 & 88.33 & 71.67 & 49.21 \\
        \texttt{QwQ-32b} ($\Gamma$) & 91.75 & 97.89 & 96.91 & 97.92 & 94.95 & 97.94 & 96.49 & 92.45 & 70.80 & 94.51 & 90.00 & 98.00 & 86.46 & 88.66 & 91.67 & 96.67 & 81.67 & 88.33 & 73.33 & 46.03 \\
        \texttt{Qwen$2.5$-$0.5$b-it} & 86.60 & 87.37 & 94.85 & 95.83 & 97.98 & 91.75 & 84.21 & 83.96 & 67.26 & 82.42 & 87.00 & 91.00 & 81.25 & 72.16 & 88.33 & 88.33 & 91.67 & 66.67 & 70.00 & 23.81 \\
        \texttt{Qwen$2.5$-$0.5$b} & 86.60 & 85.26 & 93.81 & 92.71 & 98.99 & 87.63 & 87.72 & 83.02 & 67.26 & 85.71 & 88.00 & 91.00 & 79.17 & 70.10 & 95.00 & 90.00 & 95.00 & 71.67 & 73.33 & 22.22 \\
        \texttt{Qwen$2.5$-$1.5$b-it} & 92.78 & 87.37 & 96.91 & 95.83 & 100.00 & 93.81 & 91.23 & 89.62 & 69.91 & 87.91 & 86.00 & 94.00 & 81.25 & 86.60 & 88.33 & 90.00 & 91.67 & 80.00 & 75.00 & 39.68 \\
        \texttt{Qwen$2.5$-$1.5$b} & 91.75 & 88.42 & 95.88 & 97.92 & 98.99 & 96.91 & 89.47 & 89.62 & 68.14 & 86.81 & 90.00 & 94.00 & 81.25 & 88.66 & 91.67 & 93.33 & 91.67 & 81.67 & 80.00 & 31.75 \\
        \texttt{Qwen$2.5$-14b-it} & 93.81 & 96.84 & 97.94 & 95.83 & 95.96 & 95.88 & 95.61 & 91.51 & 71.68 & 91.21 & 91.00 & 97.00 & 86.46 & 91.75 & 85.00 & 95.00 & 86.67 & 85.00 & 76.67 & 38.10 \\
        \texttt{Qwen$2.5$-14b} & 93.81 & 97.89 & 98.97 & 96.88 & 95.96 & 95.88 & 95.61 & 90.57 & 69.91 & 94.51 & 89.00 & 97.00 & 84.38 & 90.72 & 86.67 & 95.00 & 85.00 & 86.67 & 78.33 & 33.33 \\
        \texttt{Qwen$2.5$-32b-it} & 92.78 & 97.89 & 96.91 & 96.88 & 97.98 & 96.91 & 97.37 & 93.40 & 68.14 & 95.60 & 87.00 & 98.00 & 87.50 & 90.72 & 90.00 & 96.67 & 88.33 & 88.33 & 81.67 & 47.62 \\
        \texttt{Qwen$2.5$-32b} & 93.81 & 98.95 & 96.91 & 96.88 & 96.97 & 97.94 & 98.25 & 91.51 & 67.26 & 93.41 & 91.00 & 98.00 & 86.46 & 89.69 & 91.67 & 98.33 & 86.67 & 90.00 & 80.00 & 47.62 \\
        \texttt{Qwen$2.5$-3b-it} & 91.75 & 95.79 & 97.94 & 96.88 & 98.99 & 93.81 & 93.86 & 86.79 & 70.80 & 91.21 & 91.00 & 95.00 & 82.29 & 82.47 & 88.33 & 96.67 & 86.67 & 88.33 & 71.67 & 44.44 \\
        \texttt{Qwen$2.5$-3b} & 91.75 & 95.79 & 97.94 & 96.88 & 98.99 & 93.81 & 93.86 & 87.74 & 71.68 & 92.31 & 93.00 & 95.00 & 81.25 & 82.47 & 91.67 & 100.00 & 91.67 & 88.33 & 73.33 & 34.92 \\
        \texttt{Qwen$2.5$-72b-it} & 93.81 & 98.95 & 94.85 & 97.92 & 97.98 & 94.85 & 93.86 & 91.51 & 69.03 & 92.31 & 91.00 & 97.00 & 81.25 & 91.75 & 91.67 & 98.33 & 90.00 & 88.33 & 70.00 & 57.14 \\
        \texttt{Qwen$2.5$-72b} & 93.81 & 98.95 & 94.85 & 97.92 & 98.99 & 95.88 & 96.49 & 89.62 & 64.60 & 95.60 & 94.00 & 99.00 & 83.33 & 89.69 & 90.00 & 98.33 & 91.67 & 90.00 & 78.33 & 60.32 \\
        \texttt{Qwen$2.5$-7b-it} & 94.85 & 98.95 & 94.85 & 96.88 & 97.98 & 96.91 & 94.74 & 88.68 & 68.14 & 92.31 & 90.00 & 99.00 & 81.25 & 89.69 & 88.33 & 95.00 & 91.67 & 86.67 & 73.33 & 52.38 \\
        \texttt{Qwen$2.5$-7b} & 94.85 & 97.89 & 94.85 & 97.92 & 98.99 & 96.91 & 95.61 & 85.85 & 69.03 & 92.31 & 93.00 & 98.00 & 82.29 & 90.72 & 93.33 & 96.67 & 91.67 & 85.00 & 81.67 & 42.86 \\
        \texttt{Reka-flash-3} ($\Gamma$)& 93.81 & 97.89 & 92.78 & 93.75 & 93.94 & 91.75 & 83.33 & 86.79 & 69.91 & 92.31 & 91.00 & 98.00 & 86.46 & 93.81 & 86.67 & 86.67 & 76.67 & 78.33 & 71.67 & 49.21 \\
        \texttt{S1.1-32b} ($\Gamma$) & 93.81 & 98.95 & 96.91 & 96.88 & 96.97 & 96.91 & 96.49 & 92.45 & 69.03 & 93.41 & 87.00 & 98.00 & 87.50 & 88.66 & 85.00 & 98.33 & 85.00 & 91.67 & 78.33 & 49.21 \\
        \texttt{SmolLM2-$1.7$b-it} & 83.51 & 80.00 & 86.60 & 91.67 & 95.96 & 81.44 & 78.07 & 79.25 & 65.49 & 76.92 & 77.00 & 92.00 & 78.12 & 77.32 & 81.67 & 80.00 & 80.00 & 66.67 & 71.67 & 3.17 \\
        \texttt{SmolLM2-$1.7$b} & 75.26 & 81.05 & 90.72 & 91.67 & 98.99 & 84.54 & 81.58 & 79.25 & 68.14 & 80.22 & 82.00 & 93.00 & 72.92 & 71.13 & 88.33 & 78.33 & 78.33 & 63.33 & 75.00 & 1.59 \\
        \texttt{SmolLM2-135m-it} & 70.10 & 64.21 & 83.51 & 72.92 & 96.97 & 47.42 & 79.82 & 60.38 & 58.41 & 61.54 & 67.00 & 71.00 & 61.46 & 72.16 & 71.67 & 48.33 & 61.67 & 61.67 & 55.00 & 15.87 \\
        \texttt{SmolLM2-135m} & 67.01 & 68.42 & 85.57 & 79.17 & 95.96 & 54.64 & 82.46 & 59.43 & 61.06 & 63.74 & 68.00 & 78.00 & 60.42 & 58.76 & 91.67 & 66.67 & 70.00 & 53.33 & 68.33 & 1.59 \\
        \texttt{SmolLM2-360m-it} & 75.26 & 72.63 & 86.60 & 79.17 & 96.97 & 69.07 & 75.44 & 61.32 & 62.83 & 59.34 & 74.00 & 80.00 & 70.83 & 79.38 & 88.33 & 60.00 & 68.33 & 51.67 & 60.00 & 7.94 \\
        \texttt{SmolLM2-360m} & 74.23 & 72.63 & 85.57 & 84.38 & 97.98 & 71.13 & 81.58 & 61.32 & 61.95 & 65.93 & 72.00 & 85.00 & 69.79 & 75.26 & 88.33 & 65.00 & 78.33 & 55.00 & 66.67 & 9.52 \\
        \texttt{XLM-roBERTa-base} & 55.67 & 67.37 & 46.39 & 58.33 & 29.29 & 53.61 & 61.40 & 51.89 & 53.10 & 56.04 & 71.00 & 61.00 & 55.21 & 36.08 & 43.33 & 26.67 & 25.00 & 60.00 & 26.67 & 50.79 \\
        \texttt{XLM-roBERTa-large} & 63.92 & 82.11 & 79.38 & 84.38 & 68.69 & 72.16 & 75.44 & 68.87 & 64.60 & 68.13 & 82.00 & 92.00 & 68.75 & 58.76 & 76.67 & 88.33 & 73.33 & 68.33 & 56.67 & 55.56 \\
        \midrule
        \texttt{Random} & 51.55 & 49.47 & 42.27 & 53.12 & 45.45 & 37.11 & 54.39 & 53.77 & 53.10 & 47.25 & 54.00 & 55.00 & 50.00 & 49.48 & 41.67 & 51.67 & 40.00 & 53.33 & 43.33 & 55.56 \\
        \texttt{Human} & 90.72 & 90.53 & 94.85 & 90.62 & 91.92 & 92.78 & 92.98 & 89.62 & 81.42 & 86.81 & 94.00 & 93.00 & 80.21 & 98.97 & 75.00 & 90.00 & 95.00 & 73.33 & 76.67 & 84.13 \\
        \bottomrule
    \end{tabular}
    }
    \captionsetup[figure]{font=small,labelfont=small}
    \caption{The complete results average accuracy scores (\%) (higher is better) of the \llm{} LLMs and our baselines (\texttt{Random}, \texttt{Human}) by linguistic phenomena (\guillemet{LP}). \guillemet{$\Upsilon$} are model that have a specialization in French, while $\Gamma$ are model marketed as reasoning LLM. Linguistic phenomena (header) are presented in \autoref{tab:categorization_complete}.}
    \label{tab:resultscomp}
\end{table*}

\begin{table*}
    \centering
    \resizebox{\textwidth}{!}{%
    \begin{tabular}{lDDDDDDDDDDDDDDDDDDDD}
        \toprule
        LLM & \multicolumn{1}{c}{\texttt{1}} & 
        \multicolumn{1}{c}{\texttt{2}} & 
        \multicolumn{1}{c}{\texttt{3}} & 
        \multicolumn{1}{c}{\texttt{4}} & 
        \multicolumn{1}{c}{\texttt{5}} & 
        \multicolumn{1}{c}{\texttt{6}} & 
        \multicolumn{1}{c}{\texttt{7}} & 
        \multicolumn{1}{c}{\texttt{8}} &
        \multicolumn{1}{c}{\texttt{9}} & 
        \multicolumn{1}{c}{\texttt{10}} & 
        \multicolumn{1}{c}{\texttt{11}} &
        \multicolumn{1}{c}{\texttt{12}} & 
        \multicolumn{1}{c}{\texttt{13}} &
        \multicolumn{1}{c}{\texttt{14}} &
        \multicolumn{1}{c}{\texttt{15}} & 
        \multicolumn{1}{c}{\texttt{16}} & 
        \multicolumn{1}{c}{\texttt{17}} &
        \multicolumn{1}{c}{\texttt{18}} &
        \multicolumn{1}{c}{\texttt{19}} &
        \multicolumn{1}{c}{\texttt{20}} \\
        \midrule
        \texttt{Aya-23-8b} & -6.18 & -9.48 & -10.31 & 2.09 & 5.05 & -23.71 & -12.28 & -18.87 & -19.47 & -4.39 & -14.00 & -3.00 & -7.29 & -10.31 & 1.67 & -16.67 & -11.67 & -8.33 & -6.67 & -63.50 \\
        \texttt{Aya-expanse-8b} & -8.25 & -7.37 & -11.34 & -1.04 & 5.05 & -20.62 & -10.52 & -14.15 & -21.24 & 1.10 & -10.00 & 1.00 & -3.13 & -7.22 & -1.67 & -23.33 & -18.33 & -1.66 & -6.67 & -58.73 \\
        \texttt{BERT-base-French-europeana-cased} ($\Upsilon$) & -26.80 & -22.11 & -47.43 & -13.54 & -37.37 & -43.30 & -9.65 & -28.30 & -24.78 & -38.46 & -26.00 & -19.00 & -39.59 & -71.13 & 15.00 & -1.67 & -6.67 & 18.34 & -5.00 & -82.54 \\
        \texttt{Bloom-1b1} & -2.06 & 5.26 & 3.09 & 1.05 & 6.06 & 1.03 & -5.26 & -7.54 & -15.93 & 4.40 & -3.00 & 5.00 & 1.04 & -1.03 & 6.67 & 0.00 & -15.00 & 8.34 & 8.33 & -58.73 \\
        \texttt{Bloom-1b7} & -2.06 & 7.36 & 3.09 & 2.09 & 5.05 & 2.07 & -1.75 & -3.77 & -12.39 & 6.60 & -3.00 & 7.00 & 0.00 & -1.03 & 11.67 & 6.67 & -15.00 & 8.34 & 6.66 & -55.56 \\
        \texttt{Bloom-560m} & -1.03 & 0.00 & 4.12 & 2.09 & 4.04 & -1.03 & -4.38 & -3.77 & -19.47 & 4.40 & -4.00 & 4.00 & 4.17 & -1.03 & 6.67 & 0.00 & -18.33 & 3.34 & 3.33 & -63.50 \\
        \texttt{Bloom-7b1} & 1.03 & 4.21 & 2.06 & 1.05 & 4.04 & 0.00 & 0.88 & -4.71 & -12.39 & 5.50 & 1.00 & 7.00 & 5.21 & -1.03 & 5.00 & 6.67 & -11.67 & 8.34 & 6.66 & -41.27 \\
        \texttt{Bloomz-1b1} & -7.21 & 5.26 & 3.09 & 1.05 & 5.05 & 2.07 & -2.63 & -3.77 & -19.47 & 2.20 & -3.00 & 6.00 & 2.08 & -2.06 & 3.33 & -3.33 & -18.33 & 8.34 & 8.33 & -44.45 \\
        \texttt{Bloomz-560m} & -5.15 & 0.00 & -1.04 & 1.05 & 5.05 & -1.03 & -6.14 & -5.66 & -21.24 & 4.40 & -8.00 & 4.00 & 1.04 & -1.03 & 3.33 & -1.67 & -21.67 & -1.66 & 8.33 & -60.32 \\
        \texttt{CamemBERT-base} ($\Upsilon$) & -30.93 & -46.32 & -49.49 & -22.91 & -35.35 & -50.51 & -11.40 & -40.56 & -46.02 & -35.16 & -34.00 & -29.00 & -36.46 & -70.10 & 16.67 & 10.00 & -5.00 & 23.34 & -1.67 & -82.54 \\
        \texttt{CamemBERT-large} ($\Upsilon$) & -42.27 & -51.58 & -52.58 & -28.12 & -61.62 & -43.30 & -10.52 & -33.96 & -40.71 & -34.06 & -29.00 & -41.00 & -44.79 & -76.29 & 18.33 & 10.00 & -5.00 & 21.67 & -11.67 & -84.13 \\
        \texttt{Chocolatine-14b-it} ($\Upsilon$) & 0.00 & 7.36 & -1.04 & 5.21 & 4.04 & 2.07 & -0.87 & -0.94 & -5.31 & 3.30 & 0.00 & 6.00 & 4.17 & -17.53 & 11.67 & -1.67 & -6.67 & 3.34 & 1.66 & -39.69 \\
        \texttt{Chocolatine-14b-it} ($\Upsilon$) & 0.00 & 7.36 & -1.04 & 5.21 & 4.04 & 2.07 & -0.87 & -0.94 & -5.31 & 3.30 & 0.00 & 6.00 & 4.17 & -17.53 & 11.67 & -1.67 & -6.67 & 3.34 & 1.66 & -39.69 \\
        \texttt{Chocolatine-2-14b-it} ($\Upsilon$) & 1.03 & 8.42 & 1.03 & 7.30 & 5.05 & 4.13 & 2.63 & -0.94 & -9.74 & 5.50 & -4.00 & 5.00 & 4.17 & -9.28 & 8.33 & 5.00 & -13.33 & 8.34 & -3.34 & -41.27 \\
        \texttt{Claire-7b-fr-it} ($\Upsilon$) & 2.06 & 7.36 & 3.09 & 7.30 & 6.06 & 4.13 & 2.63 & 1.89 & -12.39 & 9.89 & -2.00 & 6.00 & 7.29 & -17.53 & 16.67 & 5.00 & -6.67 & 8.34 & 0.00 & -46.03 \\
        \texttt{DeepSeek-R1-distill-Llama-8b} ($\Gamma$) & 1.03 & 1.05 & 1.03 & 3.13 & 3.03 & -6.18 & -16.66 & -0.94 & -19.47 & 0.00 & -4.00 & 3.00 & 1.04 & -17.53 & 3.33 & 3.33 & -15.00 & 1.67 & -11.67 & -50.80 \\
        \texttt{DeepSeek-R1-distill-Qwen-14b} ($\Gamma$) & 1.03 & 5.26 & 3.09 & 6.26 & 6.06 & 3.10 & -6.14 & 2.83 & -12.39 & -1.10 & -1.00 & 5.00 & -3.13 & -10.31 & 8.33 & 3.33 & -11.67 & 13.34 & -3.34 & -36.51 \\
        \texttt{DeepSeek-R1-distill-Qwen-32b} ($\Gamma$) & 2.06 & 5.26 & 2.06 & 2.09 & 3.03 & 1.03 & -0.87 & -0.94 & -13.28 & 3.30 & -2.00 & 5.00 & 4.17 & -13.40 & 10.00 & 8.33 & -13.33 & 15.00 & -10.00 & -26.99 \\
        \texttt{DeepSeek-R1-distill-Qwen-7b} ($\Gamma$) & -10.31 & -10.53 & 1.03 & -4.16 & 3.03 & -7.21 & -16.66 & -16.04 & -25.67 & -15.38 & -11.00 & -5.00 & -10.42 & -17.53 & -8.33 & -5.00 & -16.67 & -11.66 & -26.67 & -47.62 \\
        \texttt{Deepthink-reasoning-14b} ($\Gamma$) & 2.06 & 6.31 & 3.09 & 4.17 & 4.04 & 3.10 & 1.76 & 1.89 & -9.74 & 5.50 & -2.00 & 5.00 & 3.12 & -8.25 & 5.00 & 3.33 & -10.00 & 13.34 & 0.00 & -46.03 \\
        \texttt{Deepthink-reasoning-7b} ($\Gamma$) & 3.09 & 8.42 & 0.00 & 7.30 & 5.05 & 4.13 & 1.76 & 0.00 & -14.16 & 4.40 & -4.00 & 6.00 & 1.04 & -10.31 & 11.67 & 3.33 & -6.67 & 13.34 & -3.34 & -30.16 \\
        \texttt{FLAN-T5-base} & -11.34 & -23.16 & -24.75 & -12.50 & -10.10 & -34.02 & -27.19 & -18.87 & -23.90 & -26.37 & -24.00 & -11.00 & -21.88 & -45.36 & -65.00 & -70.00 & -76.67 & -51.66 & -3.34 & -60.32 \\
        \texttt{FLAN-T5-large} & -12.37 & -26.32 & -31.96 & -36.45 & -2.02 & -19.58 & -14.03 & -26.41 & -35.40 & -32.96 & -16.00 & -18.00 & -27.09 & -28.87 & -56.67 & -60.00 & -75.00 & -50.00 & -10.00 & -31.75 \\
        \texttt{FLAN-T5-small} & -30.93 & -40.00 & -47.43 & -29.16 & -24.24 & -37.11 & -21.05 & -23.58 & -33.63 & -47.25 & -37.00 & -36.00 & -29.17 & -61.86 & -50.00 & -63.33 & -70.00 & -38.33 & -11.67 & -36.51 \\
        \texttt{FLAN-T5-xl} & -7.21 & -11.58 & -26.81 & -32.29 & 1.01 & -39.17 & -11.40 & -31.13 & -37.17 & -28.57 & -12.00 & -21.00 & -23.96 & -24.74 & -50.00 & -46.67 & -50.00 & -13.33 & -13.34 & -50.80 \\
        \texttt{FLAN-T5-xxl} & -16.49 & -22.11 & -42.27 & -30.20 & -12.12 & -25.77 & -19.30 & -24.53 & -30.09 & -38.46 & -17.00 & -19.00 & -27.09 & -41.24 & -20.00 & -21.67 & -50.00 & -8.33 & -13.34 & -52.38 \\
        \texttt{French-Alpaca-Llama3-8b-it} ($\Upsilon$, $\Gamma$) & 2.06 & 0.00 & 1.03 & 3.13 & 5.05 & 0.00 & -1.75 & 0.00 & -7.08 & 1.10 & -4.00 & 3.00 & 5.21 & -13.40 & 5.00 & 6.67 & -6.67 & 13.34 & -1.67 & -55.56 \\
        \texttt{Gemma-2-27b-it} ($\Gamma$) & -2.06 & -3.16 & 2.06 & 5.21 & 5.05 & -1.03 & -1.75 & -3.77 & -19.47 & 7.70 & -6.00 & 0.00 & -3.13 & -14.43 & 16.67 & 8.33 & -1.67 & 18.34 & 6.66 & -50.80 \\
        \texttt{Gemma-2-27b} ($\Gamma$) & -6.18 & 7.36 & 2.06 & 7.30 & 5.05 & 4.13 & 4.39 & -5.66 & -15.05 & 10.99 & -2.00 & 5.00 & 1.04 & -15.46 & 20.00 & 8.33 & 1.67 & 20.00 & 5.00 & -46.03 \\
        \texttt{Gemma-2-2b-it} ($\Gamma$) & -7.21 & 1.05 & 1.03 & 5.21 & 8.08 & -4.12 & -4.38 & -2.83 & -15.93 & 4.40 & -9.00 & 0.00 & 3.12 & -19.59 & 16.67 & 6.67 & -5.00 & 10.00 & -6.67 & -61.91 \\
        \texttt{Gemma-2-2b} ($\Gamma$) & -5.15 & 6.31 & 1.03 & 3.13 & 7.07 & -3.09 & -0.87 & -1.88 & -15.93 & 4.40 & -6.00 & 3.00 & -1.04 & -22.68 & 20.00 & 6.67 & -3.33 & 16.67 & -1.67 & -55.56 \\
        \texttt{Gemma-2-9b-it} ($\Gamma$) & -3.09 & 2.10 & 2.06 & 3.13 & 7.07 & -5.15 & -7.02 & -3.77 & -14.16 & 2.20 & -3.00 & 1.00 & -2.09 & -16.50 & 23.33 & 5.00 & -1.67 & 20.00 & 3.33 & -42.86 \\
        \texttt{Gemma-2-9b} ($\Gamma$) & -2.06 & 5.26 & 0.00 & 5.21 & 6.06 & 0.00 & 1.76 & 0.00 & -14.16 & 6.60 & -4.00 & 5.00 & 1.04 & -17.53 & 23.33 & 10.00 & 0.00 & 18.34 & 5.00 & -53.97 \\
        \texttt{Llama-$3.2$-1b-it} ($\Gamma$) & -3.09 & -2.11 & -2.07 & 4.17 & 7.07 & 0.00 & -10.52 & -3.77 & -12.39 & -2.19 & -6.00 & 2.00 & 2.08 & -16.50 & 1.67 & -1.67 & -8.33 & -13.33 & -10.00 & -65.08 \\
        \texttt{Llama-$3.2$-1b} ($\Gamma$) & -2.06 & -1.06 & 0.00 & 1.05 & 7.07 & -2.06 & -2.63 & -4.71 & -7.97 & 1.10 & -7.00 & 0.00 & 2.08 & -13.40 & 16.67 & 1.67 & -5.00 & 0.00 & -3.34 & -65.08 \\
        \texttt{Llama-$3.2$-3b-it}($\Gamma$) & 3.09 & 1.05 & 1.03 & 3.13 & 6.06 & 1.03 & -1.75 & -1.88 & -10.62 & 1.10 & -6.00 & 1.00 & 2.08 & -10.31 & 16.67 & 5.00 & -8.33 & 6.67 & 0.00 & -61.91 \\
        \texttt{Llama-$3.2$-3b} ($\Gamma$) & 2.06 & 3.15 & 2.06 & 3.13 & 8.08 & -1.03 & 0.88 & -1.88 & -13.28 & 6.60 & -5.00 & 3.00 & 8.33 & -9.28 & 15.00 & 5.00 & -1.67 & 20.00 & -3.34 & -63.50 \\
        \texttt{Lucie-7b-it-human-data} ($\Upsilon$) & -14.43 & -7.37 & 0.00 & -1.04 & 5.05 & 0.00 & -1.75 & -2.83 & -23.90 & 3.30 & -9.00 & 1.00 & -4.17 & -23.71 & 11.67 & 8.33 & -3.33 & 23.34 & -5.00 & -68.26 \\
        \texttt{Lucie-7b-it} ($\Upsilon$) & -11.34 & 7.36 & 1.03 & 2.09 & 8.08 & 0.00 & -0.87 & -2.83 & -15.05 & 7.70 & 0.00 & 5.00 & 1.04 & -15.46 & 21.67 & 10.00 & -3.33 & 16.67 & 1.66 & -61.91 \\
        \texttt{Lucie-7b} ($\Upsilon$) & -1.03 & 8.42 & 1.03 & 6.26 & 7.07 & 2.07 & 3.51 & 2.83 & -15.05 & 9.89 & 0.00 & 7.00 & 3.12 & -9.28 & 15.00 & 8.33 & -1.67 & 21.67 & -1.67 & -44.45 \\
        \texttt{Meta-Llama-$3.1$-70b-it} ($\Gamma$) & 1.03 & 4.21 & -1.04 & 5.21 & 5.05 & 3.10 & -3.51 & -0.94 & -10.62 & 4.40 & 1.00 & 0.00 & 3.12 & -8.25 & 20.00 & 10.00 & -6.67 & 18.34 & -3.34 & -42.86 \\
        \texttt{Meta-Llama-$3.1$-70b} ($\Gamma$) & 1.03 & 6.31 & 0.00 & 3.13 & 6.06 & 2.07 & 1.76 & 4.72 & -6.20 & 6.60 & -2.00 & 3.00 & 2.08 & -12.37 & 21.67 & 10.00 & -8.33 & 13.34 & -8.34 & -34.92 \\
        \texttt{Meta-Llama-$3.1$-8b-it} ($\Gamma$) & 1.03 & 4.21 & 2.06 & 6.26 & 6.06 & 1.03 & 0.88 & -1.88 & -2.66 & 6.60 & -2.00 & 3.00 & 4.17 & -11.34 & 15.00 & 8.33 & -3.33 & 10.00 & -5.00 & -53.97 \\
        \texttt{Meta-Llama-$3.1$-8b} ($\Gamma$)  & 0.00 & 4.21 & 3.09 & 5.21 & 5.05 & 0.00 & 1.76 & -1.88 & -7.97 & 5.50 & -5.00 & 3.00 & 2.08 & -13.40 & 10.00 & 10.00 & 0.00 & 6.67 & -1.67 & -52.38 \\
        \texttt{Ministral-8b-it} & -5.15 & 2.10 & 2.06 & 6.26 & 4.04 & 3.10 & 1.76 & -1.88 & -23.01 & 4.40 & -1.00 & 5.00 & 1.04 & -7.22 & 16.67 & 8.33 & -3.33 & 20.00 & 8.33 & -53.97 \\
        \texttt{Mistral-7b-it} & 0.00 & 4.21 & 2.06 & 3.13 & 4.04 & -3.09 & -5.26 & -0.94 & -13.28 & 4.40 & -4.00 & 3.00 & 4.17 & -19.59 & 13.33 & 5.00 & -5.00 & 18.34 & 0.00 & -57.15 \\
        \texttt{Mistral-7b} & -4.12 & 1.05 & 1.03 & 3.13 & 4.04 & 0.00 & -2.63 & -6.60 & -11.51 & 4.40 & -3.00 & 2.00 & 2.08 & -16.50 & 13.33 & 3.33 & -5.00 & 11.67 & -3.34 & -57.15 \\
        \texttt{Mistral-nemo-it} & -5.15 & 3.15 & 4.12 & 8.34 & 6.06 & 3.10 & 0.88 & 0.00 & -22.13 & 5.50 & -2.00 & 4.00 & 1.04 & -6.19 & 15.00 & 8.33 & -5.00 & 15.00 & 3.33 & -47.62 \\
        \texttt{Mistral-small-it} & 1.03 & 5.26 & 1.03 & 5.21 & 6.06 & -1.03 & 3.51 & 0.00 & -15.05 & 5.50 & 0.00 & 5.00 & 4.17 & -12.37 & 11.67 & 8.33 & -8.33 & 16.67 & 3.33 & -44.45 \\
        \texttt{Mixtral-8x7b-it} & -1.03 & 5.26 & -1.04 & 7.30 & 6.06 & 3.10 & 6.14 & 0.00 & -15.05 & 6.60 & 0.00 & 6.00 & 1.04 & -12.37 & 18.33 & 6.67 & -5.00 & 13.34 & 0.00 & -41.27 \\
        \texttt{Mixtral-8x7b-it} & 0.00 & 6.31 & -1.04 & 7.30 & 3.03 & 3.10 & 6.14 & 1.89 & -15.05 & 5.50 & 0.00 & 6.00 & 1.04 & -11.34 & 16.67 & 6.67 & -10.00 & 13.34 & 1.66 & -39.69 \\
        \texttt{Mixtral-8x7b} & -2.06 & 4.21 & 0.00 & 6.26 & 4.04 & 4.13 & 5.27 & -3.77 & -15.93 & 4.40 & -4.00 & 7.00 & 1.04 & -8.25 & 13.33 & 8.33 & -6.67 & 11.67 & -1.67 & -47.62 \\
        \texttt{Mixtral-8x7b} & 1.03 & 4.21 & 1.03 & 7.30 & 5.05 & 3.10 & 6.14 & 0.00 & -15.93 & 6.60 & -1.00 & 7.00 & 2.08 & -10.31 & 15.00 & 8.33 & -6.67 & 16.67 & 1.66 & -38.10 \\
        \texttt{Phi-3-mini-4k-it} & -2.06 & 3.15 & 2.06 & 6.26 & 4.04 & -1.03 & -3.51 & -0.94 & -12.39 & -1.10 & -2.00 & 4.00 & 2.08 & -16.50 & 10.00 & 3.33 & -11.67 & 5.00 & -3.34 & -42.86 \\
        \texttt{Phi-$3.5$-mini-it} & -2.06 & 3.15 & 3.09 & 6.26 & 7.07 & 1.03 & 0.88 & 0.00 & -7.08 & -1.10 & -4.00 & 4.00 & 3.12 & -18.56 & 15.00 & -1.67 & -5.00 & -10.00 & -6.67 & -47.62 \\
        \texttt{Phi-4} ($\Gamma$) & 2.06 & 8.42 & 2.06 & 6.26 & 5.05 & 5.16 & 4.39 & -0.94 & -9.74 & 8.79 & -1.00 & 6.00 & 2.08 & -6.19 & 8.33 & 6.67 & -8.33 & 15.00 & -5.00 & -34.92 \\
        \texttt{QwQ-32b} ($\Gamma$) & 1.03 & 7.36 & 2.06 & 7.30 & 3.03 & 5.16 & 3.51 & 2.83 & -10.62 & 7.70 & -4.00 & 5.00 & 6.25 & -10.31 & 16.67 & 6.67 & -13.33 & 15.00 & -3.34 & -38.10 \\
        \texttt{Qwen$2.5$-$0.5$b-it} & -4.12 & -3.16 & 0.00 & 5.21 & 6.06 & -1.03 & -8.77 & -5.66 & -14.16 & -4.39 & -7.00 & -2.00 & 1.04 & -26.81 & 13.33 & -1.67 & -3.33 & -6.66 & -6.67 & -60.32 \\
        \texttt{Qwen$2.5$-$0.5$b} & -4.12 & -5.27 & -1.04 & 2.09 & 7.07 & -5.15 & -5.26 & -6.60 & -14.16 & -1.10 & -6.00 & -2.00 & -1.04 & -28.87 & 20.00 & 0.00 & 0.00 & -1.66 & -3.34 & -61.91 \\
        \texttt{Qwen$2.5$-$1.5$b-it} & 2.06 & -3.16 & 2.06 & 5.21 & 8.08 & 1.03 & -1.75 & 0.00 & -11.51 & 1.10 & -8.00 & 1.00 & 1.04 & -12.37 & 13.33 & 0.00 & -3.33 & 6.67 & -1.67 & -44.45 \\
        \texttt{Qwen$2.5$-$1.5$b} & 1.03 & -2.11 & 1.03 & 7.30 & 7.07 & 4.13 & -3.51 & 0.00 & -13.28 & 0.00 & -4.00 & 1.00 & 1.04 & -10.31 & 16.67 & 3.33 & -3.33 & 8.34 & 3.33 & -52.38 \\
        \texttt{Qwen$2.5$-14b-it} & 3.09 & 6.31 & 3.09 & 5.21 & 4.04 & 3.10 & 2.63 & 1.89 & -9.74 & 4.40 & -3.00 & 4.00 & 6.25 & -7.22 & 10.00 & 5.00 & -8.33 & 11.67 & 0.00 & -46.03 \\
        \texttt{Qwen$2.5$-14b} & 3.09 & 7.36 & 4.12 & 6.26 & 4.04 & 3.10 & 2.63 & 0.95 & -11.51 & 7.70 & -5.00 & 4.00 & 4.17 & -8.25 & 11.67 & 5.00 & -10.00 & 13.34 & 1.66 & -50.80 \\
        \texttt{Qwen$2.5$-32b-it} & 2.06 & 7.36 & 2.06 & 6.26 & 6.06 & 4.13 & 4.39 & 3.78 & -13.28 & 8.79 & -7.00 & 5.00 & 7.29 & -8.25 & 15.00 & 6.67 & -6.67 & 15.00 & 5.00 & -36.51 \\
        \texttt{Qwen$2.5$-32b} & 3.09 & 8.42 & 2.06 & 6.26 & 5.05 & 5.16 & 5.27 & 1.89 & -14.16 & 6.60 & -3.00 & 5.00 & 6.25 & -9.28 & 16.67 & 8.33 & -8.33 & 16.67 & 3.33 & -36.51 \\
        \texttt{Qwen$2.5$-3b-it} & 1.03 & 5.26 & 3.09 & 6.26 & 7.07 & 1.03 & 0.88 & -2.83 & -10.62 & 4.40 & -3.00 & 2.00 & 2.08 & -16.50 & 13.33 & 6.67 & -8.33 & 15.00 & -5.00 & -39.69 \\
        \texttt{Qwen$2.5$-3b} & 1.03 & 5.26 & 3.09 & 6.26 & 7.07 & 1.03 & 0.88 & -1.88 & -9.74 & 5.50 & -1.00 & 2.00 & 1.04 & -16.50 & 16.67 & 10.00 & -3.33 & 15.00 & -3.34 & -49.21 \\
        \texttt{Qwen$2.5$-72b-it} & 3.09 & 8.42 & 0.00 & 7.30 & 6.06 & 2.07 & 0.88 & 1.89 & -12.39 & 5.50 & -3.00 & 4.00 & 1.04 & -7.22 & 16.67 & 8.33 & -5.00 & 15.00 & -6.67 & -26.99 \\
        \texttt{Qwen$2.5$-72b} & 3.09 & 8.42 & 0.00 & 7.30 & 7.07 & 3.10 & 3.51 & 0.00 & -16.82 & 8.79 & 0.00 & 6.00 & 3.12 & -9.28 & 15.00 & 8.33 & -3.33 & 16.67 & 1.66 & -23.81 \\
        \texttt{Qwen$2.5$-7b-it} & 4.13 & 8.42 & 0.00 & 6.26 & 6.06 & 4.13 & 1.76 & -0.94 & -13.28 & 5.50 & -4.00 & 6.00 & 1.04 & -9.28 & 13.33 & 5.00 & -3.33 & 13.34 & -3.34 & -31.75 \\
        \texttt{Qwen$2.5$-7b} & 4.13 & 7.36 & 0.00 & 7.30 & 7.07 & 4.13 & 2.63 & -3.77 & -12.39 & 5.50 & -1.00 & 5.00 & 2.08 & -8.25 & 18.33 & 6.67 & -3.33 & 11.67 & 5.00 & -41.27 \\
        \texttt{Reka-flash-3}( $\Gamma$) & 3.09 & 7.36 & -2.07 & 3.13 & 2.02 & -1.03 & -9.65 & -2.83 & -11.51 & 5.50 & -3.00 & 5.00 & 6.25 & -5.16 & 11.67 & -3.33 & -18.33 & 5.00 & -5.00 & -34.92 \\
        \texttt{S1.1-32b} ($\Gamma$) & 3.09 & 8.42 & 2.06 & 6.26 & 5.05 & 4.13 & 3.51 & 2.83 & -12.39 & 6.60 & -7.00 & 5.00 & 7.29 & -10.31 & 10.00 & 8.33 & -10.00 & 18.34 & 1.66 & -34.92 \\
        \texttt{SmolLM2-$1.7$b-it} & -7.21 & -10.53 & -8.25 & 1.05 & 4.04 & -11.34 & -14.91 & -10.37 & -15.93 & -9.89 & -17.00 & -1.00 & -2.09 & -21.65 & 6.67 & -10.00 & -15.00 & -6.66 & -5.00 & -80.96 \\
        \texttt{SmolLM2-$1.7$b} & -15.46 & -9.48 & -4.13 & 1.05 & 7.07 & -8.24 & -11.40 & -10.37 & -13.28 & -6.59 & -12.00 & 0.00 & -7.29 & -27.84 & 13.33 & -11.67 & -16.67 & -10.00 & -1.67 & -82.54 \\
        \texttt{SmolLM2-135m-it} & -20.62 & -26.32 & -11.34 & -17.70 & 5.05 & -45.36 & -13.16 & -29.24 & -23.01 & -25.27 & -27.00 & -22.00 & -18.75 & -26.81 & -3.33 & -41.67 & -33.33 & -11.66 & -21.67 & -68.26 \\
        \texttt{SmolLM2-135m} & -23.71 & -22.11 & -9.28 & -11.45 & 4.04 & -38.14 & -10.52 & -30.19 & -20.36 & -23.07 & -26.00 & -15.00 & -19.79 & -40.21 & 16.67 & -23.33 & -25.00 & -20.00 & -8.34 & -82.54 \\
        \texttt{SmolLM2-360m-it} & -15.46 & -17.90 & -8.25 & -11.45 & 5.05 & -23.71 & -17.54 & -28.30 & -18.59 & -27.47 & -20.00 & -13.00 & -9.38 & -19.59 & 13.33 & -30.00 & -26.67 & -21.66 & -16.67 & -76.19 \\
        \texttt{SmolLM2-360m} & -16.49 & -17.90 & -9.28 & -6.24 & 6.06 & -21.65 & -11.40 & -28.30 & -19.47 & -20.88 & -22.00 & -8.00 & -10.42 & -23.71 & 13.33 & -25.00 & -16.67 & -18.33 & -10.00 & -74.61 \\
        \texttt{XLM-roBERTa-base} & -35.05 & -23.16 & -48.46 & -32.29 & -62.63 & -39.17 & -31.58 & -37.73 & -28.32 & -30.77 & -23.00 & -32.00 & -25.00 & -62.89 & -31.67 & -63.33 & -70.00 & -13.33 & -50.00 & -33.34 \\
        \texttt{XLM-roBERTa-large} & -26.80 & -8.42 & -15.47 & -6.24 & -23.23 & -20.62 & -17.54 & -20.75 & -16.82 & -18.68 & -12.00 & -1.00 & -11.46 & -40.21 & 1.67 & -1.67 & -21.67 & -5.00 & -20.00 & -28.57 \\
        \bottomrule
    \end{tabular}
    }
    \captionsetup[figure]{font=small,labelfont=small}
    \caption{The complete results average accuracy scores (\%) (higher is better) of the \llm{} LLMs against our \texttt{human} baseline as a differential, by linguistic phenomena (\guillemet{LP}). \guillemet{$\Upsilon$} are model that have a specialization in French, while $\Gamma$ are model marketed as reasoning LLM.
    Linguistic phenomena (header) are presented in \autoref{tab:categorization_complete}.
    }
    \label{tab:resultsdiffcomp}
\end{table*}

\end{document}